\documentclass[times, review, 10pt]{elsarticle}

\usepackage{amssymb}
\usepackage{amsmath}
\usepackage{adjustbox}
\usepackage{multirow}
\usepackage{xcolor}
\usepackage{algorithm}     
\usepackage{algorithmicx}  
\usepackage{algpseudocode} 
\def\tsc#1{\csdef{#1}{\textsc{\lowercase{#1}}\xspace}}
\tsc{WGM}
\tsc{QE}
\usepackage{amsmath} 
\usepackage{amssymb} 
\usepackage{xcolor}
\usepackage{booktabs}  
\usepackage{tabularx}  
\usepackage{enumitem}
\usepackage[T1]{fontenc}
\usepackage{graphicx,verbatim}

\usepackage{adjustbox}
\usepackage{amsmath}
\usepackage{multirow}
\usepackage{booktabs}
\usepackage{tikz}
\usepackage{wasysym}
\usepackage{xcolor}
\usepackage{tabularx}
\usepackage{graphicx}  
\usepackage{fontawesome5} 
\usepackage[table]{xcolor} 
\usepackage{arydshln}   
\usepackage[colorlinks=true,
            linkcolor=blue,
            anchorcolor=blue,
            citecolor=blue,
            urlcolor=blue,
            ]{hyperref}
\usepackage{pifont}  

\usepackage{bbding}
\usepackage{amssymb}
\rowcolors{3}{gray!10}{white}
\usepackage{booktabs}
\usepackage{siunitx}
\usepackage{threeparttable}
\usepackage{xcolor}
\usepackage{caption}

\usepackage{array}
\usepackage{etoolbox}

\definecolor{bgRed}{RGB}{255, 225, 225}    
\definecolor{bgOrange}{RGB}{255, 235, 215} 
\definecolor{bgYellow}{RGB}{255, 255, 224} 
\definecolor{rowcolor}{rgb}{1.0, 0.94, 0.94}
\definecolor{mygreen}{HTML}{008080} 
\definecolor{myred}{HTML}{C71585} 
\newcommand{\githubicon}[1]{\href{#1}{\includegraphics[height=1.1em]{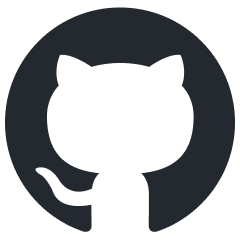}}}
\newcommand{\pul}[1]{\href{#1}{\includegraphics[height=1.1em]{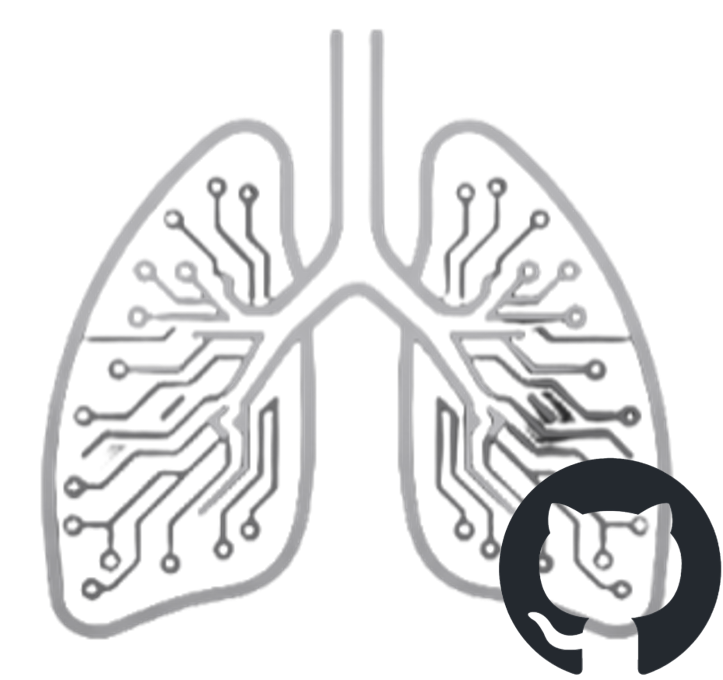}}}
\newcommand{\cmark}{{\color{mygreen}\ding{52}}}
\newcommand{\xmark}{{\color{myred}\ding{55}}}
\newcolumntype{C}{>{\centering\arraybackslash}X}
\usepackage{hyperref}

\usepackage{graphicx}
\usepackage{multirow}  

\begin{document}

\begin{frontmatter}

\title{MorVess: Morphology-Aware Pulmonary Vessel Segmentation Network}

\author{Fuyou Mao\textsuperscript{$\mathit{a}$*}}
\author{Yifei Chen\textsuperscript{$\mathit{b}$*}}
\author[hdu]{Beining Wu}
\author[csu]{Lixin Lin}
\author[hdu]{Jinnan Dai}
\author[hdu]{Zhiling Li}

\author[hdu]{Yilei Chen}
\author[hdu]{Yaqi Wang}
\author[csu]{Hao Zhang}
\author{Yan Tang\textsuperscript{$\mathit{a}$**}\protect\\[10pt]}

\author[uol]{Huiyu Zhou}
\author{Feiwei Qin\textsuperscript{$\mathit{c}$**}}

\affiliation[csu]{%
  organization={Central South University}, 
  address={410083}, 
  city={Changsha},
  country={China}}

\affiliation[thu]{%
  organization={Tsinghua University}, 
  address={100084},          
  city={Beijing},
  country={China}}

\affiliation[hdu]{%
  organization={Hangzhou Dianzi University}, 
  address={310018}, 
  city={Hangzhou},
  country={China}}

\affiliation[uol]{%
  organization={University of Leicester},
  address={LE1 7RH},
  city={Leicester},
  country={United Kingdom}}

\cortext[equal1]{These authors contributed equally to this work.}
\cortext[cor1]{Corresponding authors: Yan Tang (tangyan@csu.edu.cn), Feiwei Qin (qinfeiwei@hdu.edu.cn) .}

\begin{abstract}
Accurate pulmonary vessel segmentation remains challenging due to the sparse, tortuous, and multi-scale nature of vascular structures, where small branches are easily lost and topology integrity is difficult to preserve under voxel-wise supervision. Existing deep segmentation models primarily optimize binary masks, lacking explicit geometric constraints, thus struggling to recover continuous tubular morphology and fine vascular connectivity. In this study, we introduce MorVess, a morphology-aware segmentation framework that integrates differentiable geometric priors with large-scale foundation model adaptation to achieve fine-grained vascular parsing. MorVess jointly predicts vessel masks, distance maps, and thickness maps, providing explicit supervision for vascular boundaries, centerline consistency, and smooth diameter transitions. A lightweight 2.5D adapter bridges 3D spatial context and 2D SAM representations, while a global-local fusion block aggregates multi-level semantics and geometric cues for high-fidelity topology reconstruction. Across two challenging pulmonary CT benchmarks, MorVess delivers superior Dice, clDice, and HD95 scores, substantially improving small-vessel recovery and global connectivity. These results demonstrate that embedding geometric intelligence into pretrained vision models offers a principled and scalable pathway toward precise vessel analysis and clinically reliable structural quantification. Our source code is available at \href{https://github.com/MaoFuyou/MorVess}{https://github.com/MaoFuyou/MorVess}.

\end{abstract}



\end{frontmatter}



\section{Introduction}

Pulmonary vessel segmentation serves as a fundamental task in medical image analysis, playing a pivotal role in the diagnosis of pulmonary diseases, perfusion assessment, and preoperative planning~\cite{moccia2018blood,murugaraj2025lung}. Compared with other anatomical structures, pulmonary vessels exhibit a highly complex and sparse tree-like distribution, characterized by dense fine branches that are easily affected by noise, low contrast, and partial volume effects~\cite{deng2025mcranet,chen2024scunet++}. These challenges make it difficult to maintain both segmentation accuracy and vascular connectivity~\cite{li2025topology_2}. In particular, in the distal microvascular regions, where the vessels are exceedingly slender and tortuous, traditional methods frequently encounter issues such as discontinuities and missed detections, which severely compromise subsequent vascular quantitative analysis and functional modeling~\cite{lin2025hemorrhage,bai2025chest}.

Existing pulmonary vessel segmentation approaches can be broadly categorized into three groups: traditional image processing methods, machine learning methods based on handcrafted features, and end-to-end deep learning methods~\cite{fu2024robust}. Early studies primarily relied on techniques such as morphological operations, tubular enhancement filtering, region growing, and front propagation, often supplemented with anatomical priors to constrain vascular morphology~\cite{song2025optimized,liu2025dsdc}. However, these methods are highly susceptible to interference from lung parenchyma, particularly in regions containing fine branches and vessel crossings, making it challenging to balance recall and false detection rates. Consequently, the detection of delicate terminal structures remains suboptimal~\cite{chen2025improving}. Subsequent machine learning-based methods aimed to enhance branch recognition by incorporating manually designed multi-scale tubular and texture features. Nevertheless, their performance heavily depends on feature selection and the quality of training samples, leading to limited generalization capability~\cite{zhang2022progressive}.

With the rapid advancement of deep learning, numerous methods have been introduced in recent years to address the challenge of pulmonary vessel segmentation~\cite{wu2023transformer}. However, most of these approaches focus primarily on improving network architectures or incorporating post-processing strategies to mitigate connectivity deficiencies, yet they still struggle to fundamentally resolve topological discontinuities and geometric inconsistencies within the vascular structures~\cite{qi2023dynamic}. This issue becomes particularly pronounced under conditions of incomplete annotation or weak supervision, where models often oscillate between over-segmentation and under-segmentation. Moreover, mainstream Patch-Wise training and inference strategies disrupt the spatial correspondence between local patches and global coordinates, thereby diminishing global contextual constraints and exacerbating the risk of small-vessel disconnections~\cite{xing2024segmamba}.

Unlike the segmentation of dense structures such as tumors or organ boundaries, blood vessels represent a typical sparse topological structure, where the objective lies not only in achieving pixel-level accuracy but more importantly in maintaining global connectivity and geometric consistency\cite{wang2025serp}. Traditional deep learning models lack explicit correct geometric constraints, making it difficult for them to perceive the topological morphology and continuous spatial structure of vessels~\cite{yagis2024deep}. To overcome this limitation, this paper proposes a new paradigm for pulmonary vessel segmentation based on geometric prior constraints. Different from conventional semantic supervision based solely on binary masks, we introduce a joint supervision scheme that combines two continuously differentiable geometric prior maps tailored to tubular anatomies. These maps are jointly optimized with the semantic segmentation task to provide explicit boundary, caliber, and topological guidance for the model:

\begin{figure*}[h]
    \centering
    \includegraphics[width=1.0\linewidth]{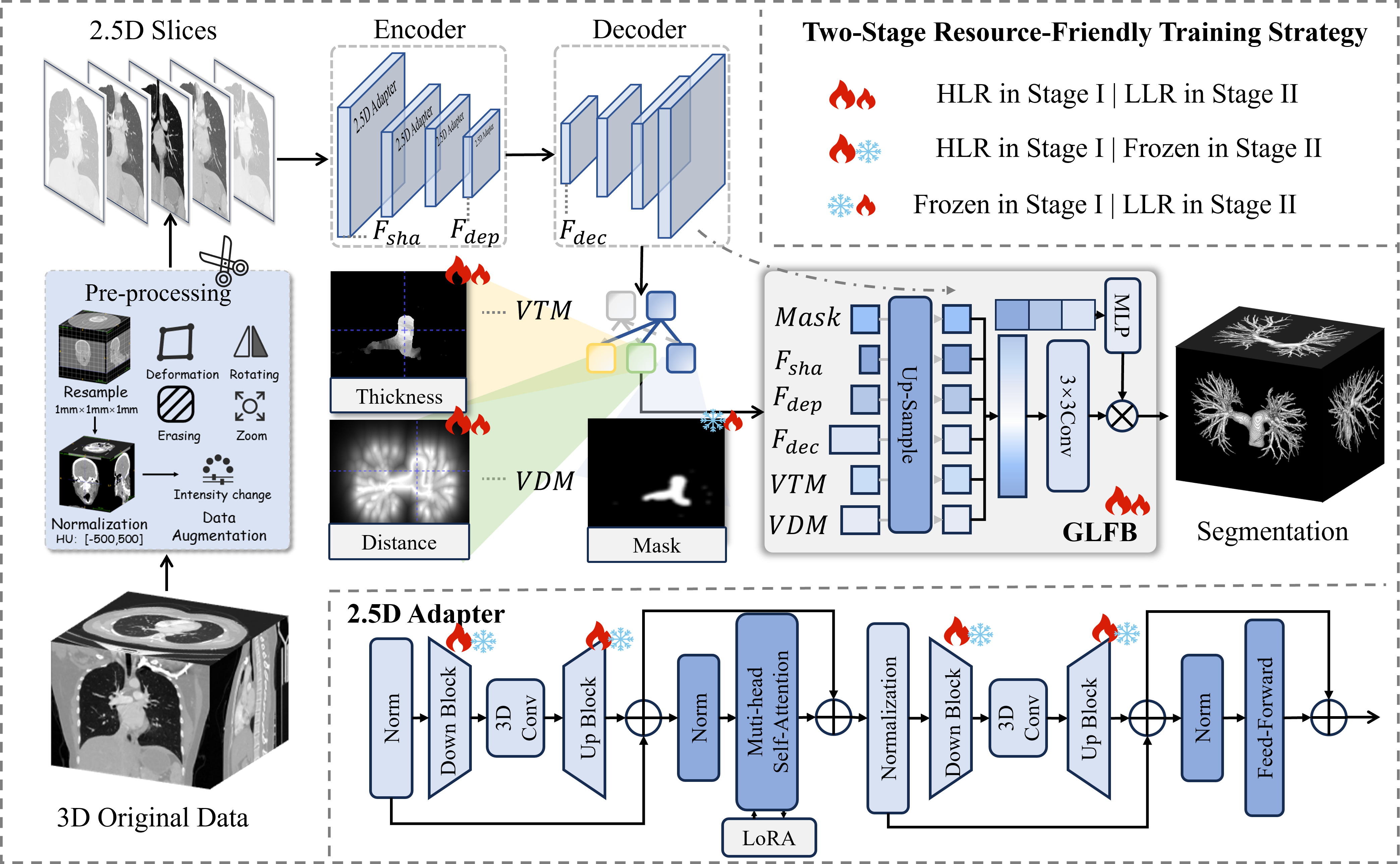}
    \caption{\normalfont \textbf{Schematic overview of the proposed MorVess framework.} A lightweight 2.5D adapter augments a frozen SAM ViT encoder with inter-slice context, while a multi-head decoder jointly predicts semantic masks and differentiable geometric priors. A global-local fusion block integrates multi-scale semantic cues with geometric fields to refine vascular topology. Two-stage fine-tuning : HLR representing Head Learning Rate for adapter initialization at $1\times{10}^{-5}$; LLR representing Late-stage Learning Rate for geometric refinement at $5\times{10}^{-5}$}.
    \label{fig:model}
\end{figure*}

In terms of model design, this paper further introduces a lightweight 2.5D hybrid segmentation framework to address the voxel-level computational burden of high-resolution CT data. By incorporating a 3D module, we efficiently adapt the 2D Segment Anything Model (SAM) to volumetric medical data, enabling the network to capture inter-slice spatial correlations while maintaining high inference efficiency. In addition, a dynamic Global Local Fusion Block (GLFB) is designed to adaptively integrate multi-scale contextual features, thereby enhancing the stability of model and coherence in fine vessel segmentation.

The main contributions are summarized as follows:
\begin{itemize}
    \item We introduce a geometry-prior segmentation that jointly optimizes VTM and VDM, substantially reinforcing the capacity to preserve vascular topology and maintain coherent geometric structure across complex branching patterns.
    \item We design a lightweight 2.5D cross-slice attention framework with an enhanced fusion block that integrates multi-level spatial cues, strengthens inter-slice coherence, and improves the continuity and completeness of fine-scale vessels.
    \item We develop a resource-efficient two-stage training strategy that enables stable and effective adaptation of large pretrained models to 3D medical segmentation, offering a practical pathway for high-fidelity vascular analysis under limited computational budgets.
\end{itemize}

\section{Related Work}
\subsection{Research on 3D Vascular Segmentation}
In three-dimensional vascular segmentation, early approaches primarily relied on image processing techniques grounded in prior anatomical knowledge. The Region Growing algorithm~\cite{almohimeed2024sandpiper} extracts vascular structures by recursively merging voxels with similar intensity values. Although conceptually straightforward, it is susceptible to the selection of initial seed points and homogeneity criteria, often leading to boundary leakage or premature termination in regions with bifurcations or pathological variations. To enhance vascular responses, researchers introduced multiscale filtering methods based on the Hessian matrix~\cite{vigneshwaran2024detection}, which quantify local tubularity through second-order derivative features to highlight vessels while suppressing background tissues. These methods perform well for large vessels but still face substantial challenges in complex vascular networks: they require manual parameter tuning, are sensitive to noise, and exhibit limited resolution in bifurcation and distal regions. Subsequent deformable models, such as Active Contour Models~\cite{chen2024automatic} and Level Set methods~\cite{lv2019vessel}, estimate vessel boundaries by evolving surfaces that minimize an energy functional, theoretically offering improved preservation of topological continuity. However, their high computational cost and strong dependence on initialization significantly constrain their applicability to clinical-scale 3D imaging.

In recent years, the rapid development of deep learning has brought a breakthrough to vascular segmentation. Researchers have continuously explored architectural innovations to better cope with the structural complexity of vascular networks. For example, Shi et al.~\cite{shi2026comma} proposed the Coordinate-Aware Network, which incorporates positional information to enhance spatial context fusion, thereby improving the recognition of fine branches; however, this method exhibits slight accuracy degradation along large vessel boundaries. Wu et al.~\cite{wu2023transformer} adopted a Transformer-based architecture and introduced a Contextual Transformer module to efficiently integrate cross-slice features, achieving remarkable performance in pulmonary artery/vein (PA/PV) segmentation tasks, though its generalization ability in distinguishing arteries and veins under small-sample conditions remains limited. Xia et al.~\cite{xia20223d} proposed ER-Net, which integrates a boundary attention module to preserve vascular wall details, leading to improved edge precision and fine-branch detection. Nevertheless, its sensitivity to vascular density and tortuosity constrains cross-dataset generalization performance.

Despite the remarkable progress achieved by deep learning methods, their core assumption still treats vascular segmentation merely as a voxel-wise classification task, lacking explicit modeling of the global topological structure \cite{luo2025pa}. Commonly used loss functions, such as cross-entropy and Dice loss, focus solely on optimizing local pixel-level correspondence, without imposing constraints on vascular connectivity or hierarchical branching relationships. As a result, these methods are prone to generating discontinuities or false connections in regions with low signal-to-noise ratios~\cite{bertels2019optimizing}. Moreover, such models are highly dependent on the quality of training annotations. When the labels contain discontinuities or incomplete regions, the network often inherits and even amplifies these annotation errors, further compromising structural consistency~\cite{wang2025smartstylemodulatedrobusttesttime}. Despite improving overall Dice via positional encoding or attention mechanisms, the aforementioned methods still fundamentally treat segmentation as independent voxel-wise classification. This overly local perspective neglects vascular topological continuity, making low-SNR terminal vessels prone to fragmentation.


\subsection{Research on SAM-based Segmentation}
SAM has attracted widespread attention in the research community due to its remarkable zero-shot segmentation capability on natural images, motivating extensive exploration of its transferability to medical image segmentation. However, the substantial differences between natural and medical images in terms of texture characteristics, contrast distribution, and semantic structure often lead to severe performance degradation when SAM is directly transferred without adaptation. Several studies have systematically quantified this domain gap: Huang et al.~\cite{huang2024segment} demonstrated under various prompt interaction settings that unadapted SAM exhibits a significant drop in Dice scores for medical tasks, while He et al.~\cite{he2023computer} reported up to a 70\% performance decline across 12 medical datasets. These findings consistently indicate that the effective deployment of SAM in medical imaging requires targeted fine-tuning and structural adaptation. Existing SAM fine-tuning methods~\cite{cheng2023sammed2d,wu2025medical}  predominantly adopt slice-wise strategies. Unfortunately, this 2D processing severs Z-axis spatial context, preventing inter-slice information utilization for tracking vascular 3D trajectories. Although recent works such as SAM-Med3D\cite{2025sam-med3d} and MedSAM-2\cite{SAM2} have begun processing entire 3D voxel data to address spatial consistency issues. However, these methods essentially still process spatial information in a patch-wise  manner, lacking explicit modeling of global 3D topological structures. This constrains their performance in recognizing fine anatomical details and assessing cross-slice connectivity. Moreover, existing SAM-based approaches predominantly rely on voxel-level mask supervision, which impedes the effective learning of tubular geometric priors from sparse annotations. Consequently, current research reveals a pronounced trade-off between prompt-driven methods that achieve high accuracy yet require labor-intensive dense annotations and lightweight fully automatic approaches that lack spatial context awareness neither of which adequately addresses the segmentation challenges of 3D tubular structures.

\subsection{Research on Vascular Topology}
In vascular segmentation tasks, preserving the integrity of the topological structure is essential for maintaining branch continuity and geometric consistency. To address this challenge, researchers have explored two primary directions: convolutional design and loss function optimization. Qi et al.~\cite{qi2023dynamic} proposed Dynamic Snake Convolution, which adaptively adjusts the convolutional kernel shape to better fit elongated and curved tubular structures, thereby enhancing the capability of network to model multiscale vascular morphology. Another line of work introduces explicit topological constraints by modifying the loss function. Shit et al.~\cite{shit2021cldice} proposed the centerline Dice (clDice) loss, which extends the supervision signal from the pixel level to the centerline level, measuring the overlap between predicted and ground truth vascular skeletons to strengthen the model’s learning of connectivity and structural consistency. Subsequently, Shi et al.~\cite{shi2024centerline} developed the centerline-boundary Dice (cbDice) loss, which augments clDice with an additional constraint on vascular boundaries, balancing centerline completeness with boundary precision. Huang et al.~\cite{huang2025harmonyseg} introduced HarmonySeg, a framework that employs a specialized loss function to balance vessel growth and noise suppression, thereby improving structural preservation under incomplete labeling conditions.

However, these approaches remain essentially post-hoc constraints applied at the binary mask level,that is, they refine or penalize topology only after segmentation results are generated. While such methods can moderately enhance connectivity, they do not provide the network with continuous and differentiable geometric supervision during training, limiting its ability to comprehend vascular spatial morphology truly. If topological constraints cannot be jointly optimized with the segmentation objective within a unified learning framework, their effects are confined to local correction rather than holistic structural reconstruction. Therefore, a challenge in current research lies in how to seamlessly integrate geometric and topological information into the segmentation learning paradigm to achieve end-to-end consistency of vascular structures.

\section{Method}

As shown in Fig. \ref{fig:model}, MorVess consists of three core modules: 1) a geometric prior map generation module, which constructs continuous geometric potential fields as supervisory signals; 2) a 2.5D geometric adaptation network, which adapts the SAM ViT encoder to three-dimensional spatial representations through a lightweight 3D adapter and enhances fine-grained vascular feature perception via a global-local fusion module; and 3) a two-stage training strategy that enables stable optimization from macroscopic structural adaptation to microscopic topological refinement.

\subsection{Generation of Geometric Priors}
To address boundary ambiguity and diameter discontinuity, we designed two physical fields: VTM (Thickness Map), based on tree-like fractal property, enforces learning of global diameter gradient patterns to prevent abrupt thinning or fragmentation; and VDM (Distance Map) transforms steep binary mask boundaries into smooth gradient potential fields, enabling distance-to-boundary perception for sub-pixel localization.

\begin{figure*}[htbp]  
    \centering
    \includegraphics[width=1.0\textwidth]{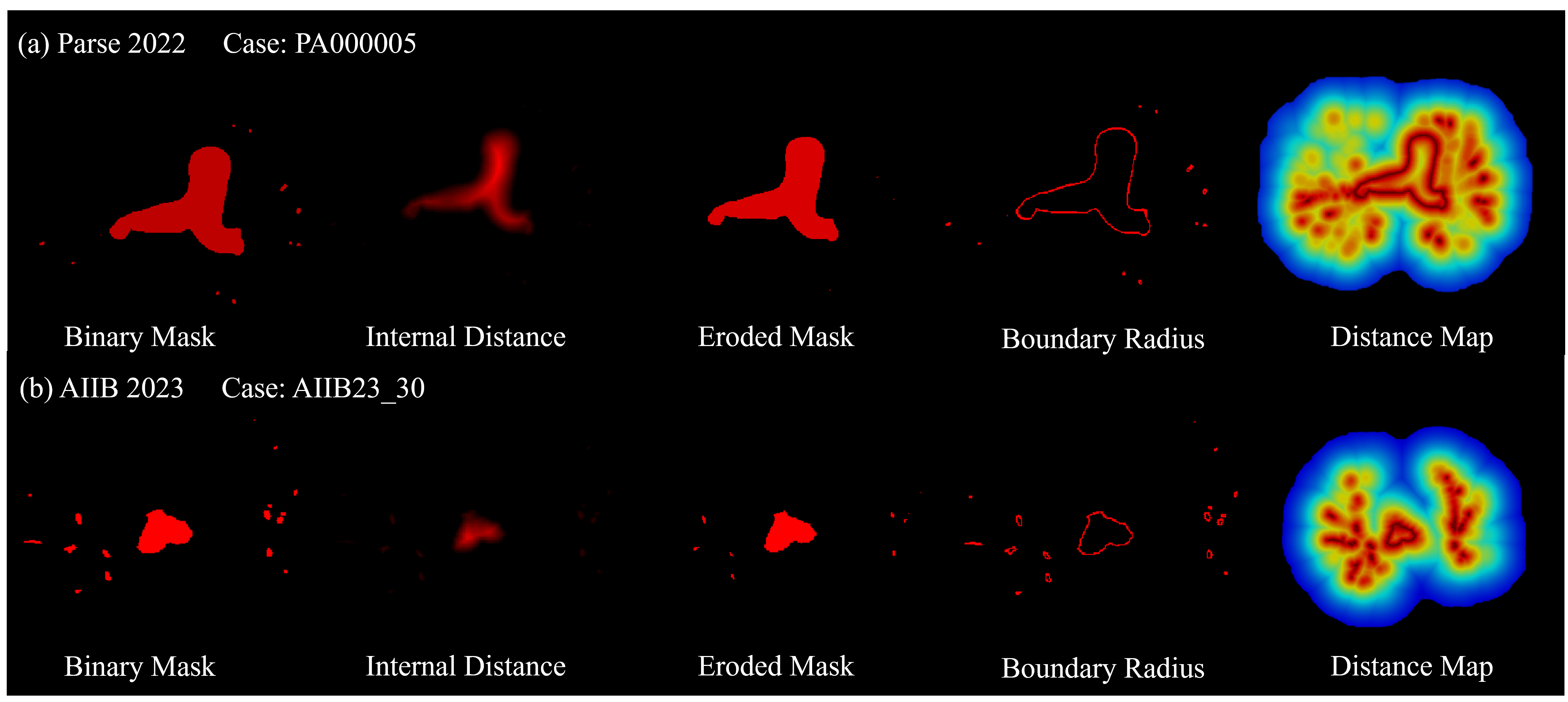} 
    \caption{\normalfont \textbf{Process of Generating Vessel Distance Map.} The process transforms discrete binary masks into continuously differentiable potential fields using morphological erosion and exponential distance decay.}
    \label{fig:Distance}
\end{figure*}

\subsubsection{Vessel Distance Map, VDM}

In standard segmentation tasks, the loss function assigns equal weights to all foreground voxels, failing to reflect the crucial role of boundary voxels in geometric structure learning. Such uniform supervision results in weak gradient signals at vessel edges, making it difficult for the model to form sharp, well-defined contours and often leading to jagged or blurred boundaries. To address this issue, we regard the vascular boundary region as a high-energy layer within the geometric potential field and construct a gradient-friendly supervisory signal that decays continuously. Fig. \ref{fig:Distance} shows the process of generating distance map from mask.

Let the discrete image space be $D \in \mathbb{Z}^3$, the set of voxels belonging to the vessel mask be $\Omega \subset D$, and the voxel spacing be represented by the vector $Sp$. First, morphological erosion is applied to $\Omega$ to obtain $\Omega \ominus \varepsilon$. The vascular boundary layer is then defined as:
\begin{equation}
\partial \Omega_{\text{gt}} = \Omega \backslash (\Omega \ominus \varepsilon),
\end{equation}
next, for any point $x \in D$, its shortest weighted Euclidean distance by SciPy, with proper handling of anisotropic voxel spacing. to the boundary layer is defined as:
\begin{equation}
d_{s_p}(x,y) = \left\| (x - y) \odot Sp \right\|_2, \quad \forall y \in \partial \Omega_{\text{gt}},
\end{equation}
the distance field is then obtained by calculating the corresponding minimum distance:
\begin{equation}
D_{\partial \Omega}(x) = \min_{y \in \partial \Omega_{\text{gt}}} d_{s_p}(x,y),
\end{equation}
finally, an exponential decay function is introduced to transform the distance field into a continuously differentiable supervisory potential field with values in $[0,1]$, referred to in this work as the VDM:
\begin{equation}
\text{VDM}(x) = \exp\left(-\lambda \cdot D_{\partial \Omega}(x)\right),
\end{equation}
where $\lambda$ is a hyperparameter controlling the decay rate, we empirically set $\lambda = 0.5$ to ensure that $\text{VDM}(x) = 1$ at the boundary and smoothly decays toward the vessel interior and exterior, thereby providing a precisely structured and reliably boundary-guided signal for model training.

\begin{algorithm}[h]
\small 
\caption{Vessel Distance Map Generation}
\label{alg:vdm}
\begin{algorithmic}[1]
\setlength{\baselineskip}{10pt}  
\setlength{\itemsep}{1pt}        

    \State \textbf{Input:} \parbox[t]{0.8\linewidth}{
        binary mask $M_{\text{gt}} \in \{0, 1\}^{D \times W \times H}$; \\
        voxel spacing vector $s=(s_z, s_y, s_x) \in \mathbb{R}^3$; \\
        decay-rate hyperparameter $\lambda$.
    }
    \State \textbf{Output:} Vessel Distance Map $M_{\text{vdm}} \in [0, 1]^{D \times W \times H}$.
    \State From the mask $M_{\text{gt}}$, extract the vascular boundary layer $M_{\text{eroded}}$ through a morphological erosion operation;
    \State Obtain the vascular boundary layer by set subtraction: $\partial\Omega_{\text{gt}} \leftarrow M_{\text{gt}} - M_{\text{eroded}}$;
    \State Initialize a three-dimensional distance field $D_{\partial\Omega}$ with the same dimensions as $M_{\text{gt}}$;
    \For{each voxel $x$ in spatial domain $D$}
        \State \parbox[t]{\dimexpr\linewidth - \algorithmicindent}{Compute the shortest Euclidean distance from voxel $x$ to the vascular boundary set $\partial\Omega_{\text{gt}}$: $D_{\partial\Omega}(x) \leftarrow d_{S_p}(x,y) = \|(x-y) \odot S_p \|_2$;}
    \EndFor
    \State Initialize the output map $M_{\text{vdm}}$ with the same dimensions as $M_{\text{gt}}$;
    \For{each voxel $x$ in spatial domain $D$}
        \State \parbox[t]{\dimexpr\linewidth - \algorithmicindent}{Transform the distance value $D_{\partial\Omega}(x)$ into a smooth potential field using an exponential function. The field equals 1 at the boundary and decays smoothly toward both the vessel interior and exterior: $M_{\text{vdm}}(x) \leftarrow \exp(-\lambda \cdot D_{\partial\Omega}(x))$;}
    \EndFor
    \State \textbf{return} $M_{\text{vdm}}$
\end{algorithmic}
\end{algorithm}

\begin{figure*}[htbp]  
    \centering  
    \includegraphics[width=1.0\textwidth]{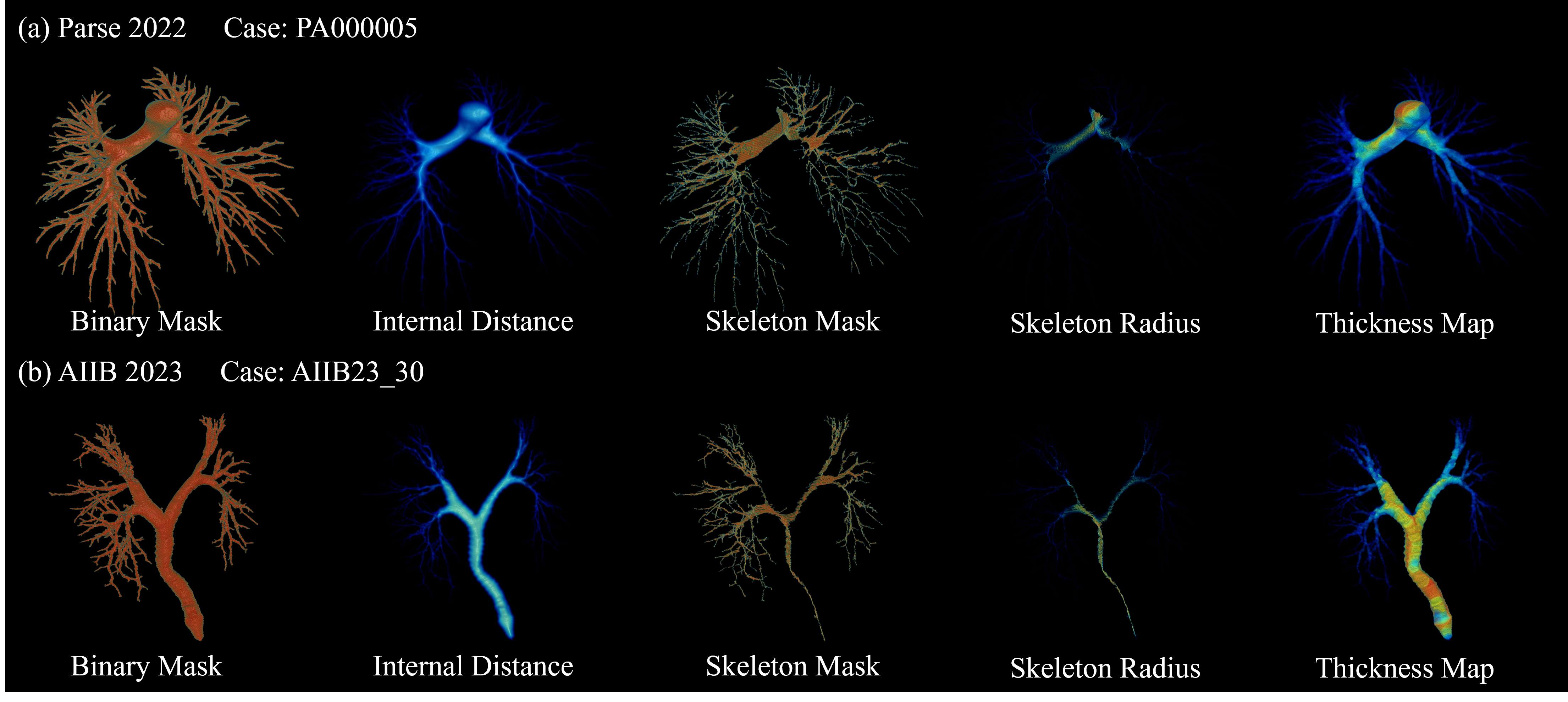}  
    \caption{\normalfont \textbf{Process of Generating Vessel Thickness Map.} The method extracts the topological skeleton from the Internal Distance field and propagates the centerline radius to the volumetric mask.}  
    \label{fig:Thickness}  
\end{figure*}

\subsubsection{Vessel Thickness Map, VTM}

The pulmonary vascular network is a three-dimensional tree-like structure characterized by global connectivity and complex branching topology. Segmentation models based on local convolution or attention mechanisms have inherently limited receptive fields, making it difficult to distinguish whether a local region belongs to a large main vessel or a fine peripheral branch. The absence of such global geometric information leads to two significant problems: (1) non-physical oscillations in the predicted vessel diameter across local regions, and (2) the loss of vascular connectivity in areas with blurred signals or low contrast. To overcome these issues, we introduce a global constraint termed the VTM, generated from the vascular centerline. VTM assigns each vascular voxel a thickness value consistent with its topological position through a medial-axis propagation algorithm, thereby providing a global-scale consistency constraint. Fig. \ref{fig:Thickness} shows the process of generating thickness map from mask.

Let the vascular region be $\Omega$, and define its internal distance field as:
\begin{equation}
D_{\text{internal}}(x) = \min_{y \in \Omega^c} d_{s_p}(x,y), \quad \forall x \in \Omega ,
\end{equation}
which represents the shortest distance from $x$ to the nearest background voxel. The vascular centerline skeleton is extracted using a topology-preserving thinning algorithm $\tau(\cdot)$  by ITK Binary Thinning Image Filter. This operator iteratively removes boundary voxels while preserving the Euler characteristic of the vascular structure $\Omega$ , ensuring that the generated skeleton $S = \tau(\Omega)$ is homotopic to the original segmentation.
Furthermore, according to the principle of the maximal inscribed sphere, we define a radius function for skeleton points, where the radius $r(s)$ of any skeleton point $s \in S$ equals its shortest distance to the vascular boundary:
\begin{equation}
r(s) = D_{\text{internal}}(s),
\end{equation}
accordingly, for any point $x$ within the vascular region $\Omega$, we use the nearest-neighbor projection operator:
\begin{equation}
\pi_s(x) = \arg\min_{s \in S} d_{s_p}(x,s),
\end{equation}
to find its corresponding point on the skeleton, and define the vascular thickness at $x$ as twice the radius of the projected skeleton point:
\begin{equation}
\text{VTM}(x) = 2 \cdot r\left(\pi_s(x)\right),
\end{equation}
thus, the final expression of the vascular thickness map is given by:
\begin{equation}
\text{VTM}(x) = 2 \cdot D_{\text{internal}}\left(\arg\min_{s \in S} d_{s_p}(x,s)\right), \quad \forall x \in \Omega,
\end{equation}
the resulting thickness map not only maintains spatial smoothness and continuity but also constrains global variations in vessel diameter, providing the model with strong geometric consistency supervision.

\begin{algorithm}[htbp]
\small 
\caption{Vessel Thickness Map Generation}
\label{alg:vtm}
\begin{algorithmic}[1]
\setlength{\baselineskip}{10pt}  
\setlength{\itemsep}{1pt}        

    \State \textbf{Input:} \parbox[t]{0.8\linewidth}{
        binary mask $M_{\text{gt}} \in \{0, 1\}^{D \times W \times H}$; \\
        voxel spacing vector $s=(s_z, s_y, s_x) \in \mathbb{R}^3$.
    }
    \State \textbf{Output:} Vessel Thickness Map $M_{\text{vtm}} \in [0, \infty)^{D \times W \times H}$.
    \State From the mask $M_{\text{gt}}$, compute the distance from each foreground voxel to the nearest background voxel using the weighted Euclidean distance transform;
    \For{each foreground voxel $x \in M_{\text{gt}}$ in the spatial domain $D$}
        \State $D_{\text{internal}}(x) \leftarrow \min_{y \notin M_{\text{gt}}} d_{s_p}(x, y)$;
    \EndFor
    \State Apply a topological ITK Binary Thinning Image Filter thinning algorithm to extract the one-dimensional centerline skeleton from the binary mask $M_{\text{gt}}$: $S_{\text{gt}} \leftarrow \text{TopologicalThinning}(M_{\text{gt}})$;
    \State Perform element-wise multiplication between the internal distance field $D_{\text{internal}}$ and the centerline skeleton $S_{\text{gt}}$ to obtain the radius value for each skeleton point: $R_s \leftarrow D_{\text{internal}} \odot S_{\text{gt}}$;
    \State For each foreground voxel $x$, find its nearest point on the skeleton $S_{\text{gt}}$;
    \For{each foreground voxel $x \in M_{\text{gt}}$ in the spatial domain $D$}
        \State $F_s(x) \leftarrow \arg\min_{s \in S_{\text{gt}}} d_{s_p}(x, s)$;
    \EndFor
    \State Initialize a zero matrix $M_{\text{vtm}}$ with the same dimensions as $M_{\text{gt}}$;
    \For{each foreground voxel $x \in M_{\text{gt}}$ in the spatial domain $D$}
        \State \parbox[t]{\dimexpr\linewidth - \algorithmicindent}{$S_{\text{nearest}} \leftarrow F_s(x)$;  // retrieve the nearest skeleton point from the index map}
        \State \parbox[t]{\dimexpr\linewidth - \algorithmicindent}{$\text{radius} \leftarrow R_s(S_{\text{nearest}})$; // obtain the radius value from the skeleton radius map}
        \State \parbox[t]{\dimexpr\linewidth - \algorithmicindent}{$M_{\text{vtm}}(x) \leftarrow 2 \times \text{radius}$; // assign the thickness value to the current voxel}
    \EndFor
    \State \textbf{return} $M_{\text{vtm}}$
\end{algorithmic}
\end{algorithm}

\subsection{Improved SAM Network Architecture}

\subsubsection{2.5D Adapter}

To fully exploit the powerful representational capacity of the 2D SAM pretrained model while simultaneously capturing the spatial contextual information inherent in 3D medical images, this study proposes an efficient 2.5D Adapter. Under the premise of keeping the SAM backbone ViT encoder frozen, this module introduces a lightweight and trainable 3D convolutional structure designed to explicitly model inter-slice depth dependencies. In doing so, it effectively bridges the gap between 2D planar features and 3D topological structures. Specifically, given an input volumetric image, the data are reconstructed along the depth dimension into a sequence of $N$ slices with the shape $[B, N, H, W]$. Before being fed into the encoder, a 2.5D Adapter module is incorporated to extract inter-slice contextual cues. This module consists of a downsampling layer, a 3D convolution layer with a kernel size of $3 \times 1 \times 1$, and an upsampling layer, which are respectively responsible for compressing feature dimensions, capturing longitudinal dependencies, and restoring channel dimensions. The adapter output is added to the original feature via a residual connection.

By fine-tuning only the Adapter parameters, the SAM encoder can naturally assimilate volumetric geometric information with negligible computational overhead, thereby markedly enhancing spatial consistency, inter-slice continuity, and topological integrity in vascular segmentation.

\subsubsection{Multi-head Geometric Prediction Decoder}

To enable the model to simultaneously perform semantic segmentation and geometric structure prediction at the output stage, this study extends the original SAM mask decoder and designs a multi-head geometric prediction decoder. On the main branch that outputs the vascular semantic mask, two additional parallel branches are introduced to regress the VDM and VTM, thereby achieving joint modeling of semantic and geometric representations.

In the original SAM decoder, a set of mask tokens interacts with the image features, and the final segmentation mask is subsequently generated through a hypernetwork. Specifically, the output mask tokens after processing by the Transformer decoder can be represented as ${h}_{\text{mask}} \in \mathbb{R}^{B * N_{\text{token}} * C}$. Each token passes through an independent multi-layer perceptron (MLP) to produce a set of dynamic weights ${W}_{\text{dyn}} \in \mathbb{R}^{B * N_{\text{token}} * C}$, these weights are then multiplied with the upsampled image features ${E}_{\text{img}}' \in \mathbb{R}^{B * C * H * W}$ to generate the initial, low-resolution segmentation mask $M_{\text{initial}}$.

Building upon this foundation, two additional dedicated geometric prediction branches are introduced, respectively responsible for generating the VDM and VTM. Unlike the main mask branch, the geometric branches do not rely on individual tokens; instead, they utilize the mean embedding of all mask tokens, defined as:
\begin{equation}
\bar{{h}}_{\text{mask}} = \text{Mean}({h}_{\text{mask}}),
\end{equation}
the global averaged embedding is fed into two independent MLPs to generate the dynamic weights ${W}_D \in \mathbb{R}^{B * N_{\text{token}} * C}$ and ${W}_T \in \mathbb{R}^{B * N_{\text{token}} * C}$.
Both sets of weights interact with the upsampled image features ${E}_{\text{img}}' \in \mathbb{R}^{B * C * H * W}$ in the same manner as the main branch, thereby generating the initial prediction logits for the two geometric priors in parallel and computationally efficiently.

\subsubsection{Global-Local Fusion Block}

Although the encoder of SAM exhibits remarkable global semantic modeling capability, its lightweight decoder relies solely on low-resolution features from the final encoder layer, causing fine-grained vascular structures such as branches and terminals to be easily overlooked. To overcome this limitation, this study designs a GLFB that integrates multi-level features and explicitly incorporates geometric priors, thereby achieving high-fidelity reconstruction of vascular structures.

The proposed module consolidates three heterogeneous yet complementary sources of information: 
(1) the feature map ${F}_{\text{dec}}$ from the standard decoder pathway, rich in global semantic context; 
(2) the shallow feature map ${F}_{\text{shallow}}$ extracted from the early encoder layers, which preserves local textures and edge details; 
(3) the deep feature map ${F}_{\text{depth}}$ from the deeper layers of encoder, providing high semantic abstraction. 

Meanwhile, to enhance vessel-specific feature learning, the self-predicted distance map VDM and thickness map VTM are incorporated as additional complementary vascular geometric priors. After being upsampled to a uniform spatial resolution, all these features are structurally concatenated along the channel dimension:
\begin{equation}
{F}_{\text{aggregated}} = \text{concat}\left[ {F}_{\text{dec}}, {F}_{\text{shallow}}, {F}_{\text{depth}}, \text{VDM}, \text{VTM} \right],
\end{equation}
the aggregated tensor is subsequently fed into a fusion subnetwork composed of $3 \times 3$ convolutional layers to promote spatial interaction among different information flows, producing a fused feature map, ${F}_{\text{fused}}$. To make the fusion process adaptive to individual samples, a lightweight dynamic convolution mechanism based on a hypernetwork is introduced. Specifically, a vessel token in the decoder of Transformer captures the global contextual representation of the current sample, denoted as ${h}_{\text{vessel}}$. This vector is then mapped through a lightweight MLP to adaptively generate a set of channel-wise attention weights:
\begin{equation}
{W}_{\text{vessel}} = \text{MLP}({h}_{\text{vessel}}),
\end{equation}
the weights are effectively applied to ${F}_{\text{fused}}$ via channel-wise weighted summation, yielding a single-channel high-resolution fusion map:
\begin{equation}
logits_{\text{vessel}} = {W}_{\text{vessel}} \cdot {F}_{\text{fused}},
\end{equation}
finally, the refined output is incorporated into the initial mask prediction through a gated residual mechanism:
\begin{equation}
M_{\text{final}} = M_{\text{initial}} + \alpha \cdot logits_{\text{vessel}},
\end{equation}
through this mechanism, the model dynamically focuses on critical vascular regions, correcting local mask prediction errors and significantly improving both fine-detailed reconstruction and topological consistency.

\subsection{Two-stage Fine-tuning Training strategy}

\paragraph{\textbf{Stage I: Macro-Feature and 2.5D Adaptation.}}
In the first stage, the model retains powerful 2D feature extraction capabilities of SAM while learning inter-slice relationships and the mapping of the newly introduced geometric prior heads. Specifically, the original ViT encoder of SAM is frozen, and only lightweight components are trained, including: (1) a 2.5D Adapter, which establishes the correspondence between 2D features and 3D voxels; and (2) a multi-head mask decoder composed of a fusion module and two output branches dedicated to predicting the distance and thickness maps. Since these modules are randomly initialized, a relatively high initial learning rate $\text{lr} = 1 \times 10^{-5}$ is adopted, combined with a warmup phase and polynomial decay scheduler, to accelerate adaptation to 3D spatial structures.

\paragraph{\textbf{Stage II: Topological Fine-tuning.}}
After completing the first stage, the trained weights are loaded for fine-tuning, aiming to enhance the perception of model of vascular boundaries, thickness, and topological connectivity. During this stage, the converged 2.5 D Adapter is frozen, and a smaller initial learning rate $\text{lr} = 5 \times 10^{-5}$ with a cosine annealing scheduler is employed to optimize the fusion module and the newly added mask prediction heads in the decoder.

\subsection{Loss function design}

To simultaneously optimize voxel-wise accuracy and topological consistency, this paper proposes a composite loss function that explicitly integrates five weighted components, achieving coherently multi-level synergistic constraints:
\begin{equation}
\mathcal{L}_{\text{total}} = \lambda_1 \mathcal{L}_{\text{CE}} + \lambda_2 \mathcal{L}_{\text{Dice}} + \lambda_3 \mathcal{L}_{\text{clDice}} + \lambda_4 \mathcal{L}_{\text{dist}} + \lambda_5 \mathcal{L}_{\text{thick}},
\end{equation}
where, $\lambda_{1-5}$ are hyperparameters controlling the relative importance of each loss term. Specifically, for the core vascular segmentation task, both cross-entropy loss ($\mathcal{L}_{\text{CE}}$) and Dice loss ($\mathcal{L}_{\text{Dice}}$) are employed to optimize voxel-level classification accuracy and regional overlap jointly: the former penalizes classification errors, while the latter alleviates class imbalance. To explicitly regularize the segmentation with respect to topological structures, a centerline Dice loss ($\mathcal{L}_{\text{clDice}}$) is introduced, measuring the consistency between predicted and ground-truth vascular skeletons and effectively suppressing breaks and false connections.

For the regression constraints of geometric prior maps, L1 loss is adopted. The vessel distance map loss ($\mathcal{L}_{\text{dist}}$) computes the L1 norm between the sigmoid-activated prediction of network $P_{\text{dist}}$ and the ground-truth distance map $G_{\text{dist}}$. For the vessel thickness map loss ($\mathcal{L}_{\text{thick}}$), a scale-invariant L1 formulation is used to ensure robustness to absolute vessel diameters. Specifically, the predicted thickness logits are first passed through a Softplus function to enforce non-negativity, yielding $P_{\text{thick}}$. Before computing the L1 loss, both $P_{\text{thick}}$ and the ground-truth $G_{\text{thick}}$ are individually and consistently normalized by their respective maximum values within each sample:
\begin{equation}
\mathcal{L}_{\text{thick}} = \left\| \frac{P_{\text{thick}}}{\max(P_{\text{thick}})} - \frac{G_{\text{thick}}}{\max(G_{\text{thick}})} \right\|_1.
\end{equation}

\section{Experiments}

\subsection{Data Acquisition}
\begin{figure}
    \centering
    \includegraphics[width=0.55\linewidth]{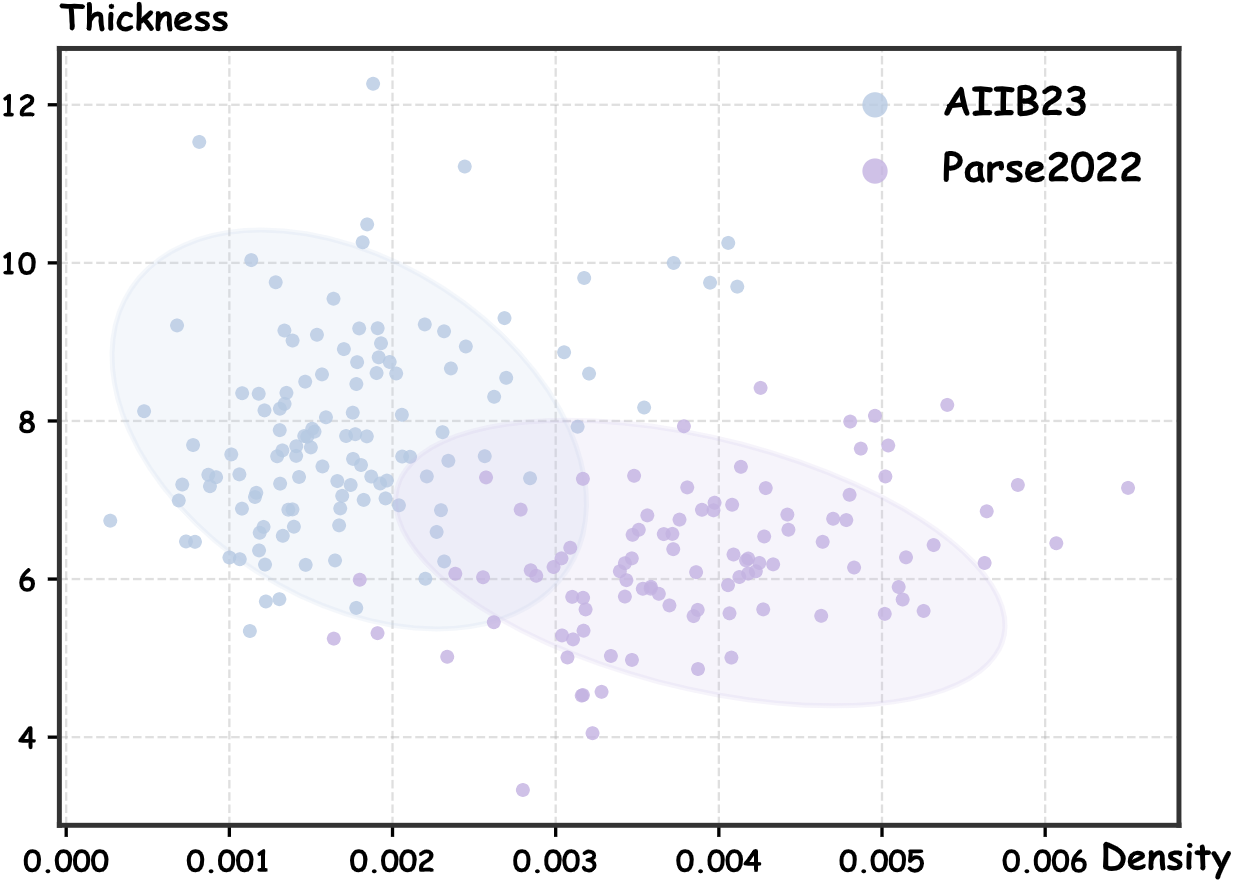}
    \caption{\normalfont \textbf{Feature Space Distribution Analysis.} The scatter plot visualizes the relationship between vascular density and thickness for the Parse2022 and AIIB23 datasets.}
    \label{fig:data_distribution}
\end{figure}

\paragraph{\textbf{Parse2022 Dataset.}}
The Parse2022 dataset used in this study originates from the MICCAI 2022 Pulmonary Artery Segmentation Challenge~\cite{maurya2022parse}. It comprises 100 3D high-resolution chest CT images along with their corresponding finely annotated segmentation ground truths. All images possess a resolution that meets clinical diagnostic standards, enabling clear visualization of multi-level vascular branching structures. As shown in Fig \ref{fig:data_distribution}, the feature space distribution is visualized.

\paragraph{\textbf{AIIB23 Dataset.}}
The Airway-Informed Imaging Biomarker 2023 (AIIB23) dataset~\cite{nan2024hunting} primarily comprises 300 chest CT scans from patients with pulmonary fibrosis, accompanied by their corresponding airway tree segmentation labels. These complex pathological morphologies not only disrupt the standard topology of vessels and airways but also introduce significant density inhomogeneity and geometric distortion, posing severe challenges to the generalization capability of automatic segmentation models. As shown in Fig \ref{fig:data_distribution}, the feature space distribution is visualized.

\paragraph{\textbf{HiPas Dataset.}}
The High-resolution Pulmonary Vessel Segmentation (HiPas) dataset \cite{chu2025deep} primarily comprises 250 high-resolution chest CT scans accompanied by corresponding pulmonary artery and vein segmentation labels. Unlike the Parse2022 dataset, which predominantly focuses on arterial structures, the HiPas dataset encompasses both arterial and venous trees, exhibiting pronounced morphological characteristics such as significant diameter gradients from central vessels to peripheral capillaries, and natural twisting and entanglement of venous trees near the left atrium.

\paragraph{\textbf{ATM2022 Dataset.}}
The Airway Tree Modeling 2022 (ATM2022) dataset \cite{zhang2023multi} primarily comprises 500 chest CT scans accompanied by corresponding airway tree segmentation labels covering the complete anatomical range from the central trachea to peripheral bronchioles. Unlike the Parse2022 dataset, which focuses on vascular structures, the ATM2022 dataset images exhibit pronounced structural characteristics in peripheral regions, such as significantly smaller diameters, more irregular branching angles, and more complex topological hierarchies.

\subsection{Data Processing}
To adapt the three-dimensional CT data to the 2.5D segmentation framework, we applied a standardized preprocessing pipeline to all datasets. First, to highlight pulmonary vascular and airway structures while suppressing background noise, the voxel intensities of all CT volumes were clipped to the range of [-500, 500] HU and linearly normalized to the intensity interval of [0, 255]. Subsequently, to incorporate local three-dimensional contextual information into the 2D network architecture, each 3D volume was sliced along the depth axis into overlapping 2D slice sequences. Specifically, for each target slice, two adjacent slices were concatenated before and after it, thereby constructing a five-channel input tensor. We perform 5-fold cross-validation and every dataset is split into five folds at the single level.

Meanwhile, during input image generation, we computed the corresponding VDM and VTM for each ground-truth mask using the vessel prior generation method. The VDM characterizes the Euclidean distance from each foreground voxel to the nearest vessel wall, thereby encoding the local boundary potential field. The VTM is derived by propagating the maximal inscribed sphere radius along the vessel centerline, reflecting the global thickness distribution.

\subsection{Experimental Settings}
All experiments were conducted on an Ubuntu 20.04 system, using the PyTorch 2.0 deep learning framework with CUDA 11.8 acceleration. The hardware setup comprised a single NVIDIA L40 GPU with 40 GB of VRAM, where both TF32 and AMP were enabled to accelerate training while maintaining numerical stability. The proposed model was built upon the SAM framework, employing ViT-Base as the encoder backbone and initializing it with the officially released pretrained weights. To achieve parameter-efficient domain adaptation, a Rank-32 Factorized Adapter Tuning (FacT) module was integrated into the encoder. Furthermore, a GLFB was designed to enhance the capture of fine-grained features. In addition, to enable geometry-aware multitask learning, two parallel decoder branches were introduced to regress the VDM and VTM, endowing the network with both topological and morphological modeling capabilities. The training process was divided into two progressive stages. The input image size was standardized to 512×512 pixels, with five adjacent slices per sample and a batch size of four. To balance pixel-wise accuracy, topological consistency, and geometric prior constraints. 
Specifically, we set $\lambda_{\text{CE}} = 0.2$, $\lambda_{\text{Dice}} = 0.8$, 
$\lambda_{\text{clDice}} = 0.3$, $\lambda_{\text{dist}} = 0.2$, and 
$\lambda_{\text{thick}} = 0.2$ for all experiments.

\textbf{Stage I: Feature Adaptation.} This phase aimed to transfer general visual representations rapidly. Only the FacT modules within the SAM encoder and the GLFB were unfrozen, and the model was trained for 400 epochs using the AdamW optimizer. The initial learning rate was set to $1\times10^{-5}$, with a warm-up period of the first 100 epochs, followed by a polynomial decay schedule with a power of 7.

\textbf{Stage II: Geometric and Topological Fine-tuning.} This phase focused on refining structural details and further enforcing topological consistency. The previously adapted FacT modules were frozen, while only the VDM and VTM decoder branches were robustly and progressively trained for 200 epochs. The AdamW optimizer was again employed with a fixed learning rate of $5\times10^{-5}$, accompanied by a cosine annealing schedule with periodic restarts to balance local exploration and global convergence.

\begin{figure*}[h!]
    \centering
    \includegraphics[width=0.95\linewidth]{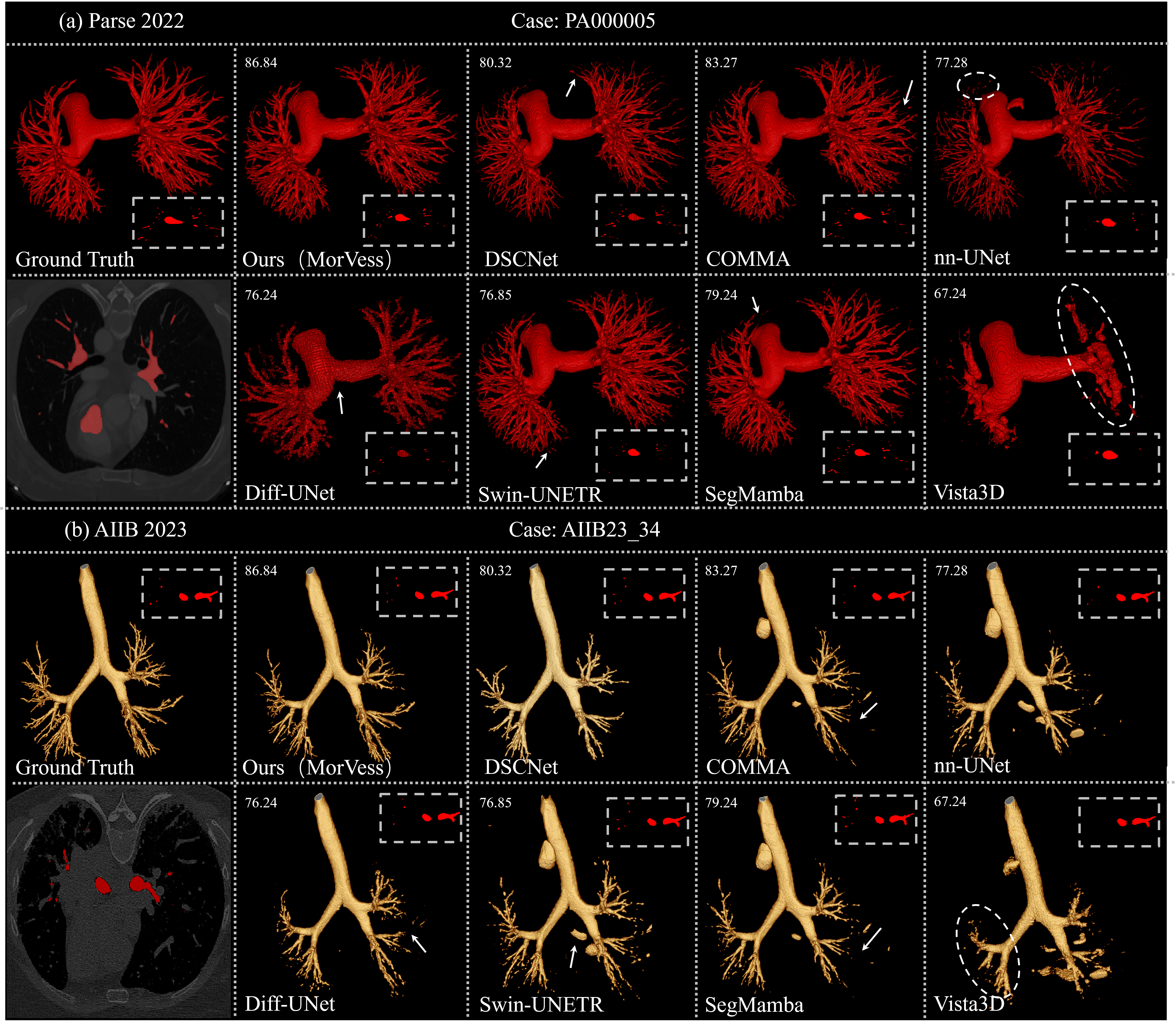}
    \caption{\normalfont \textbf{Qualitative 3D visualization of segmentation results.} The figure compares MorVess with state-of-the-art methods on (a) Parse2022 and (b) AIIB2023 datasets. White arrows and dashed circles highlight specific challenging regions.}
    \label{fig:3d_vis}
\end{figure*}

\subsection{Evaluation Indicators}
To evaluate the proposed model comprehensively, we established a multidimensional metric system covering geometric accuracy, topological connectivity, and structural integrity. The Dice Similarity Coefficient (DSC) quantifies voxel-level spatial overlap as the primary segmentation metric. The 95\% Hausdorff Distance (HD95) captures boundary-sensitive geometric conformity via the 95th percentile of maximum boundary distances. The clDice metric evaluates centerline overlap for tree-like vascular structures. The Detected Branch Ratio (DBR) and Detected Length Ratio (DLR) measure branch recovery completeness via node detection and vessel length, respectively. The Apparent Missing Rate (AMR) quantifies false negatives to assess clinical sensitivity and omission.

\subsection{Comparison Experiments}
To comprehensively assess the effectiveness of MorVess in pulmonary vessel segmentation, we conducted a systematic comparison with several representative methods on the publicly available Parse2022 and AIIB2023 datasets in Table \ref{tab:combined_results}.

First, compared to other models: nnU-Net and Swin-UNet, MorVess consistently demonstrates a significant advantage in recognizing complex tubular structures. On the Parse2022 dataset, Swin-UNet, achieves a Dice score of 76.85\%, while MorVess reaches 86.84\%, representing a relative improvement of 9.99\%. 

Compared to recent advanced vascular segmentation methods, MorVess maintains a leading position. On Parse2022, it improves Dice from 83.27\% to 86.84\% and clDice from 80.10\% to 83.22\% over the second-best comma model. Bootstrap paired resampling ($10{,}000$ iterations) confirms that clDice, DBR, and DLR all achieve $p<<0.01$ on both datasets, verifying that topological gains from geometric priors are statistically reproducible. Dice improvements reach extreme significance on Parse2022 where pulmonary CT features clear vascular boundaries and most methods approach performance ceilings and remain significant on the more challenging AIIB2023, which presents blurred boundaries and fractured micro-vessels. On the latter, geometric supervision yields pronounced improvements in HD95 and AMR. Qualitatively, 3D volumetric and 2D slice visualizations (Fig.~\ref{fig:3d_vis}) reveal that mainstream baselines suffer branch discontinuities and terminal vessel loss when recognizing small vessels, whereas MorVess preserves both quantitative accuracy and topological completeness.

\begin{table}[H]
    \centering
    \caption{\normalfont \textbf{Comparison of MorVess with other advanced methods on the Parse2022 and AIIB2023 datasets.} The \colorbox{bgRed}{best}, \colorbox{bgOrange}{second best}, and \colorbox{bgYellow}{third best} results are highlighted. $\uparrow$ indicates that higher is better, while $\downarrow$ indicates that lower is better. \textbf{Code:} \githubicon~indicates open-source code on GitHub. \textcolor{black}{The \textbf{p-value} row reports bootstrap paired resampling ($10,000$ iterations) against the second-best method}}
    \label{tab:combined_results}
    
    \normalfont
    \footnotesize
    \setlength{\tabcolsep}{2pt}
    \renewcommand{\arraystretch}{1.0}

    \newcolumntype{C}{>{\centering\arraybackslash}X}

    \begin{tabularx}{\linewidth}{l *{6}{C} c}
        \toprule
        \multicolumn{8}{c}{\textbf{Parse2022 Dataset}} \\
        \cmidrule(lr){1-8}
        \textbf{Method} & Dice$\uparrow$ & clDice$\uparrow$ & HD\textsubscript{95}(mm)$\downarrow$ & AMR$\downarrow$ & DBR$\uparrow$ & DLR$\uparrow$ & \textbf{Code} \\
        \midrule
        VISTA3D~\cite{he2024vista3d} & $78.24_{\pm 4.72}$ & $63.21_{\pm 5.97}$ & $14.23_{\pm 4.18}$ & $0.31_{\pm 0.25}$ & $0.59_{\pm 0.21}$ & $0.71_{\pm 0.15}$ & \githubicon{https://github.com/Project-MONAI/VISTA} \\
        nn-UNET-V2~\cite{liang2024automatic} & $77.28_{\pm 5.83}$ & $75.31_{\pm 5.83}$ & $9.53_{\pm 3.86}$ & $0.22_{\pm 0.22}$ & $0.66_{\pm 0.18}$ & $0.72_{\pm 0.13}$ & \githubicon{https://github.com/MIC-DKFZ/nnUNet} \\
        Swin-UNETR~\cite{hatamizadeh2021swin} & $76.85_{\pm 5.28}$ & $70.19_{\pm 5.53}$ & $11.26_{\pm 4.12}$ & $0.28_{\pm 0.23}$ & $0.61_{\pm 0.14}$ & $0.75_{\pm 0.14}$ & \githubicon{https://monai.io/research/swin-unetr} \\
        SegMamba~\cite{xing2024segmamba} & $79.24_{\pm 5.19}$ & $73.18_{\pm 4.69}$ & $9.91_{\pm 3.79}$ & $0.25_{\pm 0.14}$ & $0.65_{\pm 0.19}$ & $0.74_{\pm 0.16}$ & \githubicon{https://github.com/ge-xing/SegMamba} \\
        Diff-UNet~\cite{xing2025diff} & $76.24_{\pm 4.55}$ & $71.26_{\pm 4.27}$ & $9.65_{\pm 3.62}$ & $0.27_{\pm 0.10}$ & $0.62_{\pm 0.11}$ & $0.68_{\pm 0.11}$ & \githubicon{https://github.com/ge-xing/DiffUNet} \\
        DSCNet~\cite{hu2023dsc} & \cellcolor{bgYellow}{$80.32_{\pm 4.97}$} & \cellcolor{bgOrange}{$81.03_{\pm 3.08}$} & \cellcolor{bgYellow}{$5.35_{\pm 2.75}$} & \cellcolor{bgYellow}{$0.16_{\pm 0.15}$} & \cellcolor{bgYellow}{$0.73_{\pm 0.07}$} & \cellcolor{bgYellow}$0.78_{\pm 0.12}$ & \githubicon{https://github.com/MLMIP/DSC-Net} \\
        COMMA~\cite{shi2026comma} & \cellcolor{bgOrange}{$83.27_{\pm 4.29}$} & \cellcolor{bgYellow}{$80.10_{\pm 3.74}$} & \cellcolor{bgOrange}{$5.11_{\pm 3.42}$} & \cellcolor{bgOrange}{$0.14_{\pm 0.17}$} & \cellcolor{bgOrange}{$0.75_{\pm 0.07}$} & \cellcolor{bgOrange}{$0.79_{\pm 0.07}$} & \githubicon{https://github.com/shigen-StoneRoot/COMMA} \\
        \midrule
        MorVess(Ours) & \cellcolor{bgRed}{$86.84_{\pm 4.18}$} & \cellcolor{bgRed}{$83.22_{\pm 3.17}$} & \cellcolor{bgRed}{$4.53_{\pm 3.06}$} & \cellcolor{bgRed}{$0.12_{\pm 0.09}$} & \cellcolor{bgRed}{$0.80_{\pm 0.08}$} & \cellcolor{bgRed}{$0.83_{\pm 0.08}$} & \pul{https://github.com/MaoFuyou/MorVess.git} \\
        \midrule
        \multicolumn{1}{l}{\textit{\textbf{p-value}}} & \multicolumn{1}{C}{\footnotesize $\mathbf{2.5\times10^{-4}}$} & \multicolumn{1}{C}{\footnotesize $\mathbf{<10^{-4}}$} & \multicolumn{1}{C}{\footnotesize $0.119$} & \multicolumn{1}{C}{\footnotesize $0.253$} & \multicolumn{1}{C}{\footnotesize $\mathbf{<10^{-4}}$} & \multicolumn{1}{C}{\footnotesize $\mathbf{1.0\times10^{-4}}$} & \\
        \bottomrule
    \end{tabularx}

    \vspace{4pt} 

    \begin{tabularx}{\linewidth}{l *{6}{C} c}
        \multicolumn{8}{c}{\textbf{AIIB2023 Dataset}} \\
        \cmidrule(lr){1-8}
        \textbf{Method} & Dice$\uparrow$ & clDice$\uparrow$ & HD\textsubscript{95}(mm)$\downarrow$ & AMR$\downarrow$ & DBR$\uparrow$ & DLR$\uparrow$ & \textbf{Code} \\
        \midrule
        VISTA3D~\cite{he2024vista3d} & $83.81_{\pm 9.81}$ & $76.24_{\pm 5.22}$ & $9.23_{\pm 6.23}$ & $0.25_{\pm 0.15}$ & $0.58_{\pm 0.19}$ & $0.67_{\pm 0.14}$ & \githubicon{https://github.com/Project-MONAI/VISTA} \\
        nn-UNET-V2~\cite{liang2024automatic} & \cellcolor{bgYellow}{$92.83_{\pm 6.55}$} & $84.31_{\pm 5.09}$ & $5.92_{\pm 6.01}$ & \cellcolor{bgYellow}{$0.10_{\pm 0.14}$} & $0.77_{\pm 0.14}$ & $0.83_{\pm 0.12}$ & \githubicon{https://github.com/MIC-DKFZ/nnUNet} \\
        Swin-UNETR~\cite{hatamizadeh2021swin} & $90.53_{\pm 9.81}$ & $80.13_{\pm 4.88}$ & $8.42_{\pm 5.82}$ & $0.19_{\pm 0.12}$ & $0.74_{\pm 0.15}$ & $0.78_{\pm 0.10}$ & \githubicon{https://monai.io/research/swin-unetr} \\
        SegMamba~\cite{xing2024segmamba} & $91.29_{\pm 7.24}$ & $85.51_{\pm 5.29}$ & \cellcolor{bgYellow}{$4.59_{\pm 6.11}$} & $0.12_{\pm 0.17}$ & $0.68_{\pm 0.13}$ & $0.79_{\pm 0.15}$ & \githubicon{https://github.com/ge-xing/SegMamba} \\
        Diff-UNet~\cite{xing2025diff} & $90.48_{\pm 7.19}$ & \cellcolor{bgYellow}{$86.32_{\pm 4.42}$} & $4.67_{\pm 5.57}$ & $0.11_{\pm 0.08}$ & $0.64_{\pm 0.16}$ & $0.76_{\pm 0.06}$ & \githubicon{https://github.com/ge-xing/DiffUNet} \\
        DSCNet~\cite{hu2023dsc} & $92.15_{\pm 5.84}$ & $85.22_{\pm 4.74}$ & $5.39_{\pm 5.23}$ & \cellcolor{bgYellow}{$0.10_{\pm 0.12}$} & \cellcolor{bgOrange}{$0.80_{\pm 0.12}$} & \cellcolor{bgYellow}$0.81_{\pm 0.08}$ & \githubicon{https://github.com/MLMIP/DSC-Net} \\
        COMMA~\cite{shi2026comma} & \cellcolor{bgOrange}{$92.88_{\pm 5.25}$} & \cellcolor{bgOrange}{$86.23_{\pm 3.94}$} & \cellcolor{bgOrange}{$4.25_{\pm 4.94}$} & \cellcolor{bgOrange}{$0.09_{\pm 0.02}$} & \cellcolor{bgYellow}{$0.81_{\pm 0.13}$} & \cellcolor{bgOrange}{$0.84_{\pm 0.11}$} & \githubicon{https://github.com/shigen-StoneRoot/COMMA} \\
        \midrule
        MorVess(Ours) & \cellcolor{bgRed}{$94.31_{\pm 3.52}$} & \cellcolor{bgRed}{$89.34_{\pm 3.46}$} & \cellcolor{bgRed}{$3.24_{\pm 4.81}$} & \cellcolor{bgRed}{$0.07_{\pm 0.04}$} & \cellcolor{bgRed}{$0.86_{\pm 0.09}$} & \cellcolor{bgRed}{$0.89_{\pm 0.16}$} & \pul{https://github.com/MaoFuyou/MorVess.git} \\
        \midrule
        \multicolumn{1}{l}{\textit{\textbf{p-value}}} & \multicolumn{1}{C}{\footnotesize $\mathbf{1.9\times10^{-2}}$} & \multicolumn{1}{C}{\footnotesize $\mathbf{<10^{-4}}$} & \multicolumn{1}{C}{\footnotesize $\mathbf{1.1\times10^{-2}}$} & \multicolumn{1}{C}{\footnotesize $\mathbf{8.0\times10^{-4}}$} & \multicolumn{1}{C}{\footnotesize $\mathbf{<10^{-4}}$} & \multicolumn{1}{C}{\footnotesize $\mathbf{2.1\times10^{-2}}$} & \\
        \bottomrule
    \end{tabularx}
\end{table}

\subsection{Ablation Experiments}
\subsubsection{Effectiveness of Geometric-prior supervision}

\begin{figure}
    \centering
    \includegraphics[width=1.0\linewidth]{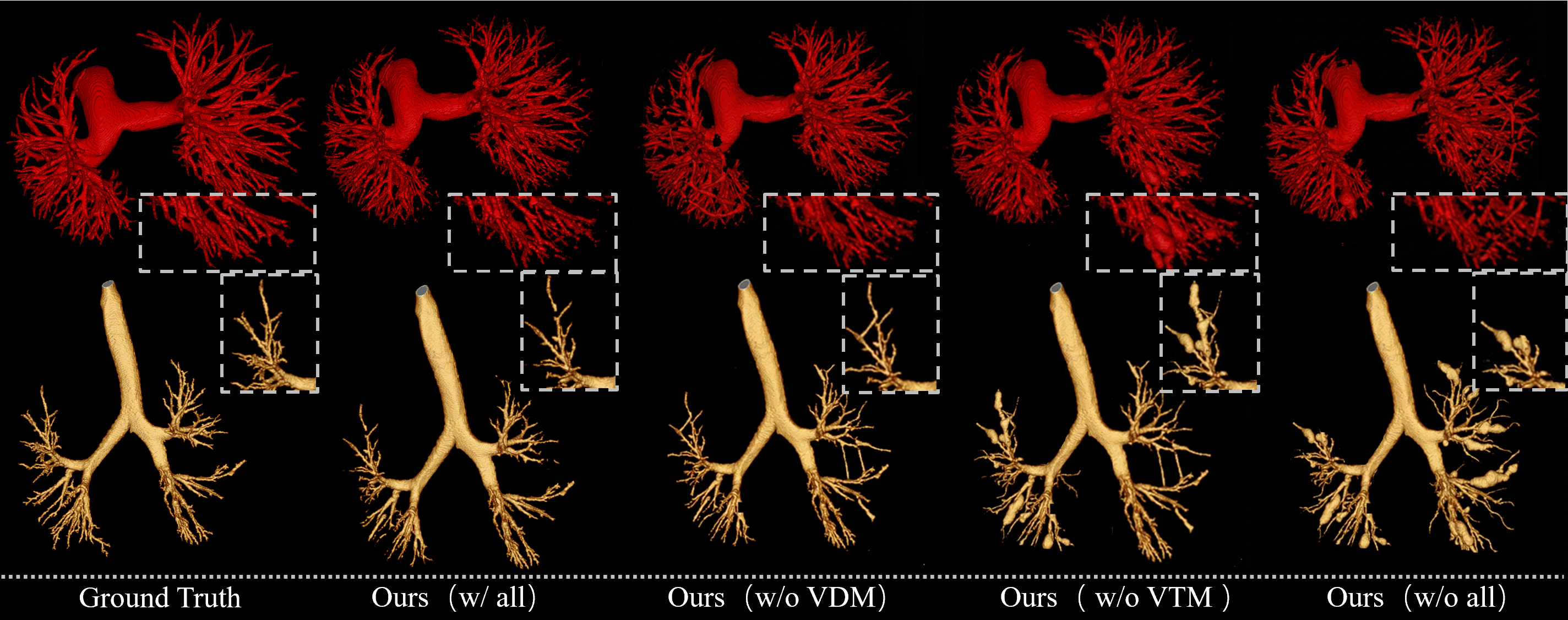}
    \caption{\textbf{Impact of the Two Geometric Priors on Segmentation Performance.} Qualitative comparison on Parse2022 and AIIB2023 datasets. The zoomed-in insets demonstrate that the full model preserves fine distal branches and smooth boundaries, while removing either prior leads to discontinuities or diameter oscillations in peripheral regions.}
    \label{fig:placeholder}
\end{figure}

To validate the impact of the geometric prior supervision framework, we designed a stepwise ablation experiment in Table \ref{tab:ablation_geometry} and visualization comparsion in Figure \ref{fig:placeholder}. Starting from a baseline that only receives standard binary-mask supervision within our 2.5-D architecture, the introduction of distance map supervision lowers HD95 from 6.85 mm to 5.72 mm, visibly reducing discontinuities in thin branches and improving lumen continuity.
Incorporating thickness map supervision further lifts clDice from 74.24\% to 78.75\%, substantially suppressing boundary thickness variations and segmentation artifacts.
Under the full geometric prior, both metrics improve synergistically, confirming that explicit anatomical constraints effectively alleviate blurred vascular boundaries and unstable predictions of fine branches.

\begin{table}[htbp]
\centering
\caption{\normalfont \textbf{Ablation study on geometric-prior supervision conducted on the Parse2022 dataset.} Mean Dice, clDice, HD95 (mm), AMR, DBR, and DLR are reported, with the best results highlighted in bold.}
\label{tab:ablation_geometry}
\normalfont
\small 
\setlength{\tabcolsep}{4pt}
\definecolor{rowcolor}{rgb}{1.0, 0.94, 0.94}
\newcolumntype{L}{>{\raggedright\arraybackslash}X}
\newcolumntype{C}{>{\centering\arraybackslash}X}

\begin{tabularx}{\linewidth}{ >{\cellcolor{rowcolor}}L *{6}{C} }
\toprule
\multicolumn{7}{c}{\textbf{Parse2022 Dataset}} \\
\cmidrule(lr){1-7}
\textbf{Method} & Dice$\uparrow$ & clDice$\uparrow$ & HD\textsubscript{95}(mm)$\downarrow$ & AMR$\downarrow$ & DBR$\uparrow$ & DLR$\uparrow$ \\
\midrule
Baseline &
$82.4_{\textcolor{myred}{\pm 5.89}}$ &
$74.24_{\textcolor{myred}{\pm 4.15}}$ &
$6.85_{\textcolor{myred}{\pm 5.22}}$ &
$0.22_{\textcolor{myred}{\pm 0.15}}$ &
$0.62_{\textcolor{myred}{\pm 0.14}}$ &
$0.72_{\textcolor{myred}{\pm 0.16}}$ \\
$+$VDM &
$83.9_{\textcolor{myred}{\pm 4.53}}$ &
$79.32_{\textcolor{myred}{\pm 3.86}}$ &
$5.72_{\textcolor{myred}{\pm 4.86}}$ &
$0.18_{\textcolor{myred}{\pm 0.13}}$ &
$0.68_{\textcolor{myred}{\pm 0.11}}$ &
$0.78_{\textcolor{myred}{\pm 0.13}}$ \\
$+$VTM &
$84.1_{\textcolor{myred}{\pm 4.65}}$ &
$78.75_{\textcolor{myred}{\pm 4.02}}$ &
$5.59_{\textcolor{myred}{\pm 4.52}}$ &
$0.19_{\textcolor{myred}{\pm 0.10}}$ &
$0.67_{\textcolor{myred}{\pm 0.12}}$ &
$0.77_{\textcolor{myred}{\pm 0.09}}$ \\
\scriptsize{$+$VDM$+$VTM} &
$\textbf{86.84}_{\textcolor{myred}{\pm \textbf{4.18}}}$ &
$\textbf{83.22}_{\textcolor{myred}{\pm \textbf{3.17}}}$ &
$\textbf{4.53}_{\textcolor{myred}{\pm \textbf{3.06}}}$ &
$\textbf{0.12}_{\textcolor{myred}{\pm \textbf{0.09}}}$ &
$\textbf{0.80}_{\textcolor{myred}{\pm \textbf{0.08}}}$ &
$\textbf{0.83}_{\textcolor{myred}{\pm \textbf{0.08}}}$ \\
\bottomrule
\end{tabularx}
\end{table}

\subsubsection{Effectiveness of each key component}
Table~\ref{tab:ablation_components} ablates the 2.5D Adapter, GLFB, and SAM pretraining. The baseline achieves a Dice of 0.6844. Adding GLFB alone improves it to 0.7481, confirming the importance of cross-scale semantic fusion. The 2.5D Adapter alone raises Dice to 0.7233, verifying that inter-slice contextual modeling is crucial for 3D vessel continuity. Their combination yields 0.7626, demonstrating complementary effects. SAM pretraining further boosts performance: with the 2.5D Adapter alone, Dice reaches 0.8392, highlighting the value of strong visual priors. The complete model achieves 0.8544, validating the essential role of each component.

\begin{table}[h!]
\centering
\caption{\small \textbf{Ablation study on the contribution of different model components.} The checkmark (\cmark) indicates the component is used, while the cross (\xmark) indicates it is not. The best performance is highlighted in bold.}
\label{tab:ablation_components}

\small
\renewcommand{\arraystretch}{0.85}  
\setlength{\tabcolsep}{8pt}        

\begin{tabular}{ccc c}
\toprule
\textbf{Pretrained} & \textbf{2.5D} & \textbf{Global-Local} & \textbf{Dice Score} $\uparrow$ \\
\textbf{Weights}    & \textbf{Adapter} & \textbf{Block} & \\
\midrule
\xmark & \xmark & \xmark & 0.6844 \\
\xmark & \cmark & \xmark & $0.7233_{\textcolor{myred}{+0.0389}}$ \\
\xmark & \xmark & \cmark & $0.7481_{\textcolor{myred}{+0.0637}}$ \\
\xmark & \cmark & \cmark & $0.7626_{\textcolor{myred}{+0.0782}}$ \\
\cmark & \xmark & \cmark & $0.8033_{\textcolor{myred}{+0.1189}}$ \\
\cmark & \cmark & \xmark & $0.8392_{\textcolor{myred}{+0.1548}}$ \\
\cmark & \cmark & \cmark & $\textbf{0.8544}_{\textcolor{myred}{\textbf{+0.1700}}}$ \\
\bottomrule
\end{tabular}
\end{table}

\begin{figure}
    \centering
    \includegraphics[width=0.4\linewidth]{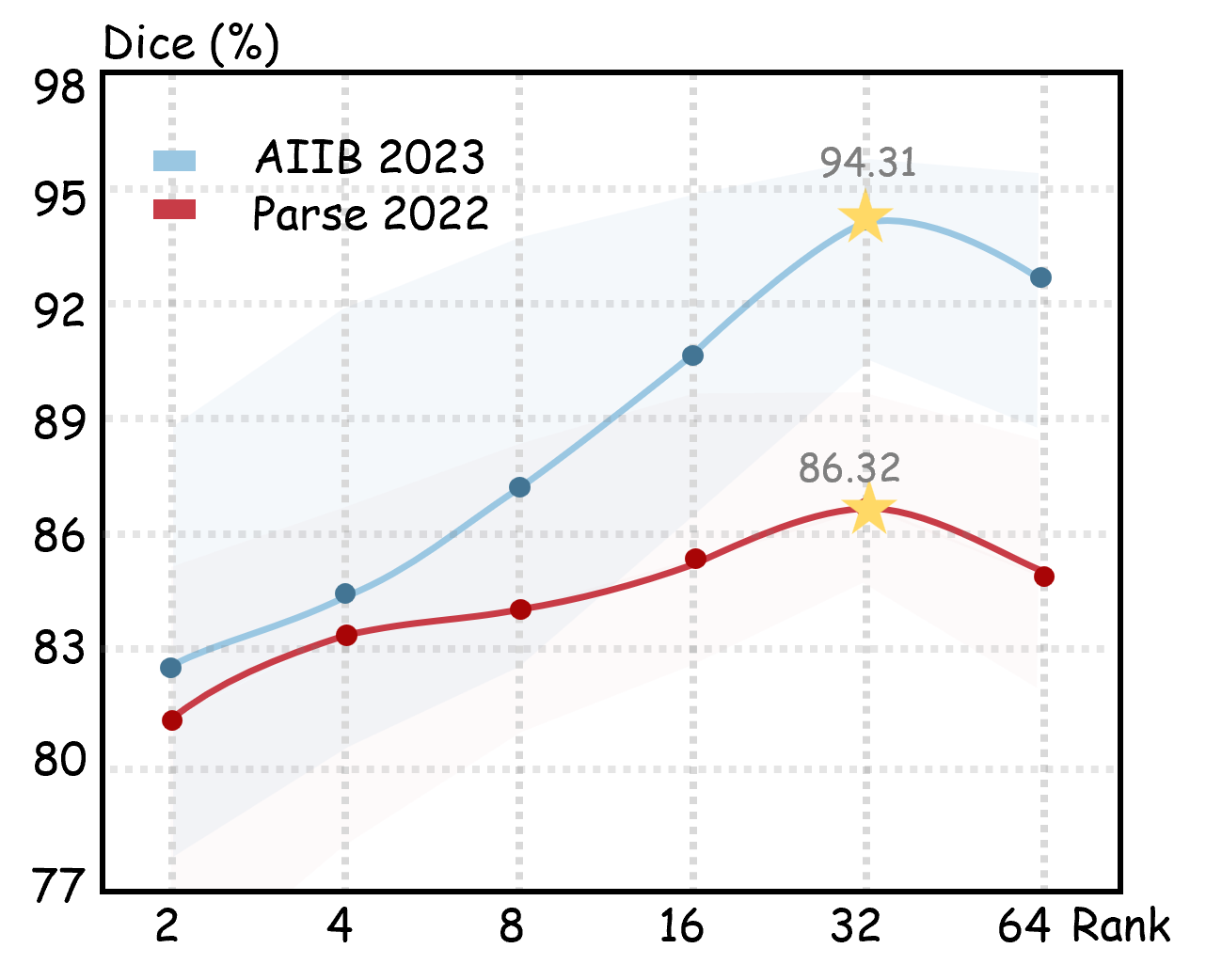}
    \caption{\normalfont \textbf{Impact of LoRA/FaCT rank on segmentation accuracy.} Dice scores steadily increase with larger ranks and gradually saturate around rank 32, indicating an overall optimal balance between representational capacity and parameter efficiency.}
    \label{fig:abliation-3}
\end{figure}

\subsubsection{Effectiveness of different choice about the rank}
To balance performance and computational cost, we evaluated the model with low-rank decomposition ranks $r$ set to $\{2, 4, 8, 16, 32, 64\}$. The results, as shown in Fig.~\ref{fig:abliation-3}, indicate that the model performance gradually improves as the rank increases. However, when $r = 32$, the performance gain over $r = 16$ is less than 1\%, suggesting that the performance has reached saturation. Therefore, we ultimately selected $r = 32$ as the optimal trade-off, achieving a balance between performance enhancement and the introduction of additional parameters.

\subsubsection{Effectiveness on the computational efficiency and resource cost}

To evaluate the computational efficiency and resource demands of our model, we conducted a comparative analysis against baseline methods, with the results detailed in Table \ref{tab:efficiency}. The comparison highlights the significant resource-friendliness of MorVess. MorVess requires only 1.0 M trainable parameters, which is substantially lower than nnU-Net and Diff-UNet. In terms of computational load, our method consumes only 42 GMACs per 5-slice stack, a fraction of the 180 GMACs required by nnU-Net and 340 GMACs by Diff-UNet. Furthermore, the peak memory usage is drastically reduced, with MorVess utilizing only 4.2 GB of VRAM, compared to 18 GB for nnU-Net and 32 GB for Diff-UNet. The ablation results demonstrate that these modules add minimal computational cost and negligible memory footprint . This confirms that the performance gains attributed to these modules are achieved with exceptional efficiency.

\begin{table}[h]
\centering
\small
\caption{\small \textbf{Efficiency comparison on Parse2022 dataset.} We report trainable/total params (M), GMACs/slice stack, peak memory (GB@batch=4), and iterations to reach 95\% of the final Dice.}
\label{tab:efficiency}
\normalfont
\renewcommand{\tabularxcolumn}[1]{m{#1}}
\setlength{\tabcolsep}{3pt} 
\begin{tabularx}{\linewidth}{l C C C C C C}
\toprule
\multicolumn{1}{c}{{\small \textbf{Method}}} & {\small \textbf{Trained param (M)}} & {\footnotesize \textbf{Total params (M)}} & {\small \textbf{GMACs/ slice stack}} & {\small \textbf{Volume/s}} & {\footnotesize \textbf{GB}} & {\small \textbf{Epoch /iters}} \\
\midrule
nnU-Net & 32 & 32 & 180 & 0.35 & 18 & {\scriptsize 600/15 k} \\ 
Diff-UNet & 64 & 64 & 340 & 0.18 & 32 & {\scriptsize 900/22.5 k} \\
\addlinespace[2pt]
\cdashline{1-7}
\addlinespace[3pt]
w/o GLFB & 0.55 & 93.2 & 39 & 1.75 & 3.9 & 380/9.5k \\
w/o 2.5D Adapter & 0.92 & 93.5 & 40 & 1.70 & 4.1 & 420/10.5k \\
\textcolor{myred}{MorVess (Ours)} & \textcolor{myred}{1.0} & \textcolor{myred}{93.6} & \textcolor{myred}{42} & \textcolor{myred}{1.6} & \textcolor{myred}{4.2} & \textcolor{myred}{450/11k} \\
\bottomrule
\end{tabularx}
\end{table}

\subsection{Cross-dataset analysis}

{To validate the robustness of MorVess against vascular geometric variations simulating challenges such as vascular tortuosity and abnormal diameters under pathological conditions,
we performed cross-domain generalization tests using two out-of-distribution datasets, HiPas and ATM2022.

We evaluate cross-domain generalization by directly applying the Parse2022 trained model to the HiPas venous subset and the AIIB2023 trained model to the peripheral ATM2022 test set. On HiPas, MorVess achieves a Dice of 81.14\% and retains the global venous topology despite arterial-to-venous intensity shifts; as shown in Figure~\ref{fig:ood}, it preserves continuity even in tortuous left-atrium regions. This confirms that VTM-based thickness supervision captures tubular invariants beyond intensity appearance. On ATM2022, the model attains a Dice of 89.25\% and a clDice of 86.75\% under a dual domain shift, decreasing only slightly from the in-domain 94.31\%. These results demonstrate that VDM/VTM constraints encode topological continuity rather than dataset-specific photometric cues, generalizing across anatomical systems.

\begin{table}[t]
\centering
\caption{\textbf{Cross-dataset generalization results.} The MorVess model trained on the source domain is directly evaluated on the target domain without any fine-tuning. Best results in each group are highlighted in bold.}
\label{tab:cross_dataset}
\resizebox{\textwidth}{!}{%
\begin{tabular}{@{}lllcccccc@{}}
\toprule
\textbf{Experiment} & \textbf{Train Domain} & \textbf{Test Domain} & \textbf{Dice$\uparrow$} & \textbf{clDice$\uparrow$} & \textbf{HD95$\downarrow$} & \textbf{AMR$\downarrow$} & \textbf{DBR$\uparrow$} & \textbf{DLR$\uparrow$} \\
\midrule
In-Domain   & Parse2022 & Parse2022 & \textbf{86.84$\pm$4.18} & \textbf{83.22$\pm$3.17} & \textbf{4.53$\pm$3.06} & \textbf{0.12$\pm$0.09} & \textbf{0.80$\pm$0.08} & \textbf{0.83$\pm$0.08} \\
Cross-Domain& Parse2022 & HiPas & 81.14$\pm$3.58 & 78.42$\pm$4.20 & 7.18$\pm$2.12 & 0.19$\pm$0.07 & 0.73$\pm$0.08 & 0.76$\pm$0.07 \\
\midrule
In-Domain   & AIIB2023  & AIIB2023  & \textbf{94.31$\pm$3.52} & \textbf{89.34$\pm$3.46} & \textbf{3.24$\pm$4.81} & \textbf{0.07$\pm$0.04} & \textbf{0.86$\pm$0.09} & \textbf{0.89$\pm$0.16} \\
Cross-Domain& AIIB2023  & ATM2022   & 89.25$\pm$2.45 & 86.75$\pm$3.10 & 4.22$\pm$1.30 & 0.10$\pm$0.04 & 0.83$\pm$0.05 & 0.86$\pm$0.04 \\
\bottomrule
\end{tabular}%
}
\end{table}

\begin{figure}
    \centering
    \includegraphics[width=1.05\linewidth]{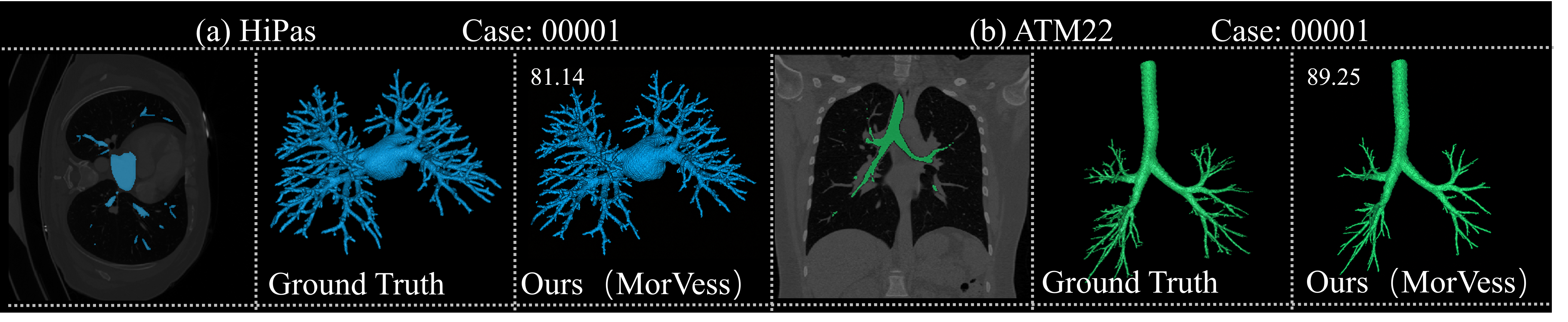}
    \caption{\textbf{Qualitative 3D visualization of cross-dataset results on HiPas and ATM2022.} The MorVess model trained solely on the source domain is directly evaluated on the target domain without fine-tuning.}
    \label{fig:ood}
\end{figure}

\subsection{Geometric Analysis}
To validate the reliability of the model in maintaining vascular tree structures from an anatomical perspective, we used the Vascular Modeling Toolkit (VMTK) to extract the centerlines of the model-predicted (MP) and ground-truth (GT) vascular segmentations, and then evaluated geometric consistency from three dimensions:

\begin{figure*}[h]  
    \centering  
    \includegraphics[width=1.0\textwidth]{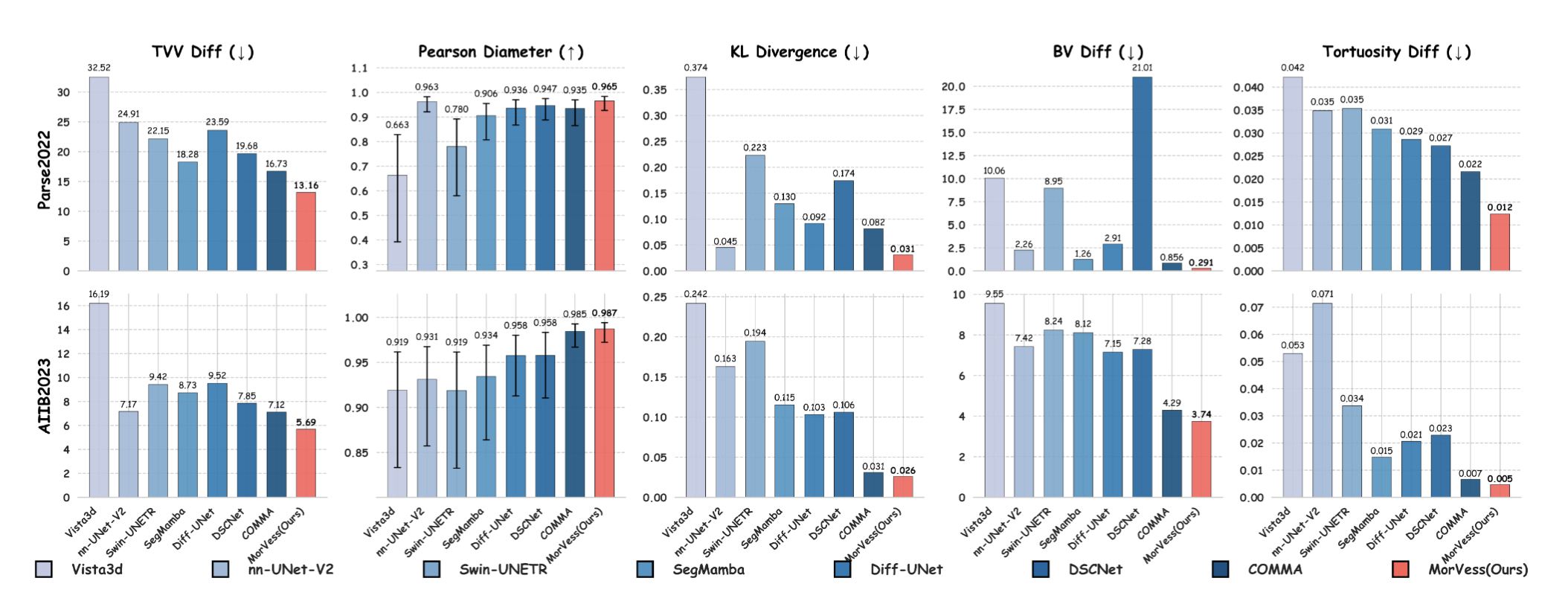}  
    \caption{\normalfont \textbf{Geometric consistency analysis across domains.} The top and bottom rows correspond to Parse2022 and AIIB2023 datasets, respectively.}  
    \label{fig:geometric}  
\end{figure*}
\subsubsection{TVV Consistency}
Total vessel volume (TVV) is a key physiological parameter that reflects the overall scale  of tubular structures. Accurate estimation of this parameter is crucial for assessing disease burden and treatment response. As shown in Fig.~\ref{fig:geometric},
on Parse2022, MorVess yields the lowest TVV difference of 13.16, surpassing the second-best COMMA and clearly below all other baselines. On AIIB2023, MorVess again attains the lowest TVV difference percentage versus 7.12\% for COMMA. This indicates that MorVess exhibits extremely low systematic error and high consistency in estimating vascular volume, ensuring high clinical reliability. TVV is associated with increased pulmonary blood flow and is routinely used as a non-invasive indicator for pulmonary arterial hypertension screening and left heart related pulmonary venous congestion, whereas reduced TVV may reflect chronic thromboembolic disease or advanced interstitial lung disease with vascular pruning. 

\subsubsection{Diameter Distribution Consistency}
Vessel diameter is a physiological parameter reflecting the morphology of the vessel wall. We analyzed the diameters of all sampled points along the centerline and plotted the distribution histogram. As shown in Fig \ref{fig:geometric}, 
MorVess achieves the highest Pearson correlation between predicted and GT diameter distributions on both datasets. KL divergence is lowest on Parse2022 and competitive on AIIB2023. The high Pearson correlation and low KL divergence achieved by MorVess indicate that the predicted vessel tree preserves the physiologic caliber gradient from central elastic arteries to peripheral muscular arterioles. This fidelity is particularly important for emerging computational hemodynamic analyses that rely on accurate lumen geometry to non-invasively stratify PH severity without dedicated CT angiography\cite{TELOKEN2026109285}.

\subsubsection{Consistency of Small-Vessel Volume Fraction}
The small-vessel volume fraction is a key indicator reflecting the ability of the model to preserve small branches. Following the definition in ~\cite{shi2026comma}, we stratified the vascular system based on cross-sectional area and calculated the volume fraction of structures with a cross-sectional area smaller than 5 mm\textsuperscript{2}. As shown in Fig \ref{fig:geometric}, for small-vessel fidelity, MorVess exhibits lowest BV difference on Parse2022 and low Tortuosity difference, indicating superior distal-branch preservation with correct curvature trends. On AIIB2023, MorVess maintains very low errors though COMMA is slightly lower. Combined with MorVess leading TVV and Pearson, these outcomes suggest a balanced geometric reliability: strong small-branch recovery without over-smoothing Tortuosity. Loss of small vessels is one of the earliest and most sensitive imaging signatures of pulmonary vascular remodeling in interstitial lung disease (ILD) and COPD\cite{SPAGNOLO2025739}.

\section{Limitation}

The 2.5D adapter intentionally trades a fraction of geometric fidelity for substantial gains in computational efficiency and GPU memory. This design choice, however, brings two practical constraints: the Z-axis receptive field is capped at $\pm$2 slices, which limits the capture of long range anatomical dependencies spanning 30-50 mm, and the implicit quasi-isotropic assumption makes the model less robust to anisotropic voxels common in clinical CT. To address these limitations, future work will pursue hybrid attention mechanisms that extend the Z-axis context without inflating the number of parameters, as well as multi-site validation across diverse CT acquisition protocols.

\section{Conclusion}
In this study, we proposed MorVess, a morphology-aware framework that couples differentiable vascular priors with SAM-based adaptation to achieve anatomically faithful pulmonary vessel segmentation. By jointly modeling masks, distance maps, and thickness maps, MorVess enforces explicit topological and morphological constraints, enabling reliable recovery of fine distal branches and continuous vascular trees. Extensive experiments demonstrate consistent gains in Dice, clDice, and HD95 across challenging datasets, highlighting the value of embedding geometric intelligence into pretrained vision models.

\section*{CRediT authorship contribution statement}

\textbf{Fuyou Mao}: Methodology, Visualization, Writing and original draft.  
\textbf{Yifei Chen}: Conceptualization, Methodology, Writing and original draft.  
\textbf{Beining Wu}: Validation, Visualization.  
\textbf{Lixin Lin}: Validation.  
\textbf{Jinnan Dai}: Visualization.  
\textbf{Zhiling Li}: Visualization.  
\textbf{Yilei Chen}: Validation.  
\textbf{Yaqi Wang}: Writing and review \& editing.  
\textbf{Hao Zhang}: Supervision, Writing and review \& editing. 
\textbf{Yan Tang}: Writing and review \& editing.  
\textbf{Huiyu Zhou}: Writing and review \& editing.  
\textbf{Feiwei Qin}: Project administration, Writing and review \& editing.

\section*{Declaration of competing interest}
The authors declare that they have no known competing financial interests or personal relationships that could have influenced the work reported in this paper.

\section*{Acknowledgments}
This work was supported by the High Performance Computing Center of Central South University, Fundamental Research Funds for the Provincial Universities of Zhejiang (No. GK259909299001-006), the State Key Lab of CAD\&CG, Zhejiang University (A2510), Anhui Provincial Joint Construction Key Laboratory of Intelligent Education Equipment and Technology (No. IEET202401), and the Graduate Research and Innovation Project of Central South University (No. 1053320241117).

\bibliographystyle{elsarticle-num-names}
\bibliography{ref}
\end{document}